\PassOptionsToPackage{table}{xcolor}
\documentclass[lettersize,journal]{IEEEtran}

\usepackage{amsmath,amsfonts}
\usepackage{algorithm}
\usepackage{algorithmic}
\usepackage{graphicx}
\usepackage{booktabs}
\usepackage{multirow}
\usepackage{makecell}
\usepackage{xcolor}
\usepackage{arydshln}
\usepackage{pifont}
\usepackage{colortbl}
\usepackage{enumitem}
\usepackage{balance}
\usepackage[caption=false,font=footnotesize]{subfig}

\usepackage[most]{tcolorbox}
\tcbuselibrary{breakable,skins}
\usepackage[framemethod=tikz]{mdframed}
\usepackage{framed}
\usepackage{alltt}
\usepackage{fontawesome5}
\usepackage{adjustbox}
\usepackage{listings}
\usepackage{threeparttable}
\usepackage{textcomp}
\usepackage{stfloats}
\usepackage{url}
\usepackage{verbatim}
\usepackage{cite}

\usepackage[table]{xcolor}

\newcommand{\cA}[1]{\cellcolor{blue!38}#1}  
\newcommand{\cB}[1]{\cellcolor{blue!28}#1}  
\newcommand{\cC}[1]{\cellcolor{blue!12}#1}  
   
\definecolor{ablrow}{gray}{0.90}

\definecolor{promptbg}{gray}{0.95}
\colorlet{shadecolor}{promptbg}
\definecolor{myblue}{RGB}{26,111,212}
\definecolor{mybluelight}{RGB}{235,243,255}
\definecolor{mybluetitle}{RGB}{232,242,255}

\newtcolorbox{promptbox}[1][]{
  enhanced jigsaw,
  breakable,
  colback=blue!3!white,
  colframe=blue!50!black,
  coltitle=white,
  fonttitle=\bfseries\small,
  boxrule=0.8pt,
  arc=2pt,
  left=6pt,
  right=6pt,
  top=4pt,
  bottom=4pt,
  title=#1,
  before upper={\linespread{0.92}\selectfont}
}

\newcommand{\fit}[1]{\adjustbox{max width=\linewidth}{#1}}

\newcommand{\method}{UniCVR}
\newcommand{\ie}{\textit{i.e.}}
\newcommand{\eg}{\textit{e.g.}}

\begin{document}

\title{UniCVR: From Alignment to Reranking for Unified Zero-Shot Composed Visual Retrieval}

\author{
Haokun Wen,
Xuemeng Song\IEEEmembership{, Senior Member, IEEE},
Haoyu Zhang,
Weili Guan\IEEEmembership{, Member, IEEE},
Xiangyu Zhao\IEEEmembership{, Member, IEEE},
and Liqiang Nie\IEEEmembership{, Senior Member, IEEE}%
\thanks{Haokun Wen is with the School of Computer Science and Technology, Harbin Institute of Technology (Shenzhen), Shenzhen 518055, China, and also with the Department of Data Science, City University of Hong Kong, Hong Kong 999077, China. E-mail: haokunwen@outlook.com.}
\thanks{Xuemeng Song is with the Department of Computer Science and Engineering, Southern University of Science and Technology, Shenzhen 518055, China. E-mail: sxmustc@gmail.com.}%
\thanks{Haoyu Zhang is with the School of Computer Science and Technology, Harbin Institute of Technology (Shenzhen), Shenzhen 518055, China, and also with Pengcheng Laboratory, Shenzhen 518000, China. E-mail: zhang.hy.2019@gmail.com.}%
\thanks{Weili Guan is with the School of Information Science and Technology, Harbin Institute of Technology (Shenzhen), Shenzhen 518055, China, and also with Shenzhen Loop Area Institute, Shenzhen 518045, China. E-mail: honeyguan@gmail.com.}%
\thanks{Xiangyu Zhao is with the Department of Data Science, City University of Hong Kong, Hong Kong 999077, China. E-mail: xianzhao@cityu.edu.hk.}%
\thanks{Liqiang Nie is with the School of Computer Science and Technology, Harbin Institute of Technology (Shenzhen), Shenzhen 518055, China. E-mail: nieliqiang@gmail.com.}%
\thanks{Xuemeng Song, Xiangyu Zhao, and Liqiang Nie are the corresponding authors.}%
}

\markboth{}%
{Wen \MakeLowercase{\textit{et al.}}: UniCVR: From Alignment to Reranking for Unified Zero-Shot Composed Visual Retrieval}

\maketitle

\begin{abstract}
Composed image retrieval, multi-turn composed image retrieval, and composed video retrieval all share a common paradigm: composing the reference visual with modification text to retrieve the desired target. Despite this shared structure, the three tasks have been studied in isolation, with no prior work proposing a unified framework, let alone a zero-shot solution. In this paper, we propose UniCVR, the first unified zero-shot composed visual retrieval framework that jointly addresses all three tasks without any task-specific human-annotated data. UniCVR strategically combines two complementary strengths: Multimodal Large Language Models (MLLMs) for compositional query understanding and Vision-Language Pre-trained (VLP) models for structured visual retrieval. Concretely, UniCVR operates in two stages. In Stage~I, we train the MLLM as a compositional query embedder via contrastive learning on a curated multi-source dataset of approximately 3.5M samples, bridging the heterogeneous embedding spaces between the MLLM and the frozen VLP gallery encoder. A cluster-based hard negative sampling strategy is proposed to strengthen contrastive supervision. In Stage~II, we introduce an MLLM-guided dual-level reranking mechanism that applies adaptive budgeted subset scoring to a small number of top-ranked candidates, and then exploits the resulting relevance signals through a dual-level re-scoring scheme, producing more accurate final rankings with minimal computational overhead. Extensive experiments across five benchmarks covering all three tasks demonstrate that UniCVR achieves cutting-edge performance, validating its effectiveness and generalizability.
\end{abstract}


\section{Introduction}
\label{sec:intro}

\begin{figure}[htp]
    \centering
	\includegraphics[width=0.95\linewidth]{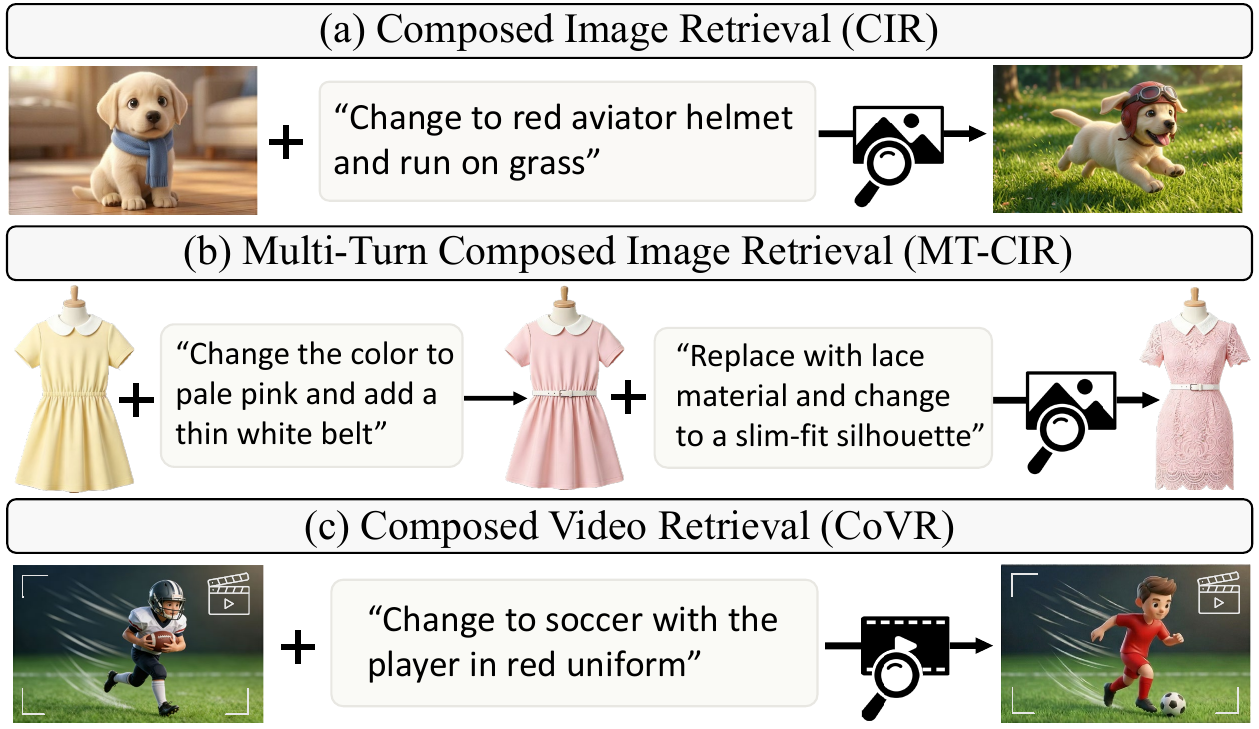}
	\caption{Illustration of Composed Visual Retrieval.}
	\label{fig:cvr_task}
\end{figure}

Visual retrieval has evolved beyond simple text or image queries. Users increasingly want to specify \emph{what to change} about a reference visual, composing it with natural language modification intent to retrieve the intended target. This need manifests across diverse scenarios, including composed image retrieval (CIR)~\cite{tgcir,offset,limn,survey}, multi-turn composed image retrieval (MT-CIR)~\cite{cfir,dialog,fashionntm}, and composed video retrieval (CoVR)~\cite{covr,covr2,hud}, where the reference visual (image or video) is paired with modification texts to retrieve the intended target, as illustrated in Figure~\ref{fig:cvr_task}.

Despite their surface differences, these tasks share two core requirements: \textbf{multimodal query understanding}, which demands jointly comprehending the reference visual and modification text to accurately capture the intended target, and \textbf{cross-modal visual retrieval}, which requires matching the composed query against a visual gallery within a well-structured metric space. To date, no existing work has attempted to unify all three tasks under a single framework. This observation motivates us to develop a unified approach for composed visual retrieval (CVR) tasks. Considering that collecting annotated data for supervised training is prohibitively expensive and often domain-specific, we focus on the \emph{zero-shot} setting, where no human-labeled data is available, making this approach both necessary and practical.

Although no prior work studies all three tasks under a unified zero-shot setting, existing studies on individual tasks provide valuable insights. Among them, ZS-CIR is by far the most mature setting~\cite{survey}, whereas MT-CIR has mainly been investigated under supervised learning~\cite{cfir,dialog,fashionntm} and zero-shot CoVR remains largely unexplored~\cite{covr,covr2}. Since the three tasks share the same core requirements, the limitations of existing ZS-CIR methods, the most developed of the three, serve as a useful lens on the challenges confronting any unified CVR framework.
In essence, most of these ZS-CIR methods reduce composed retrieval to conventional text-to-image retrieval, \ie, the multimodal query is first converted into a textual form, which is then encoded by the text encoder of a Vision-Language Pre-trained (VLP) model (\eg, CLIP~\cite{clip}) and matched against the gallery. They differ mainly in how this conversion is performed.
\textbf{Textual-inversion-based methods}~\cite{pic2word,searle,fti4cir} train a lightweight network to map the reference image into one or more pseudo-word embeddings in the word space of the VLP model, which are concatenated with the modification text and jointly encoded by the VLP text encoder. 
However, compressing the visual query into a few pseudo tokens inevitably discards fine-grained visual information, limiting the fidelity of the multimodal query representation~\cite{fti4cir,survey}. Moreover, the remaining query understanding burden falls entirely on the VLP text encoder, which is optimized for short descriptive captions rather than multimodal compositional reasoning, inherently capping retrieval performance. These limitations become even more pronounced for richer inputs, such as videos with spatio-temporal dynamics and multi-turn queries with accumulated interaction history.

\textbf{Training-free (M)LLM-based methods}~\cite{seize,ldre,autocir} instead offload query understanding to Large Language Models (MLLMs/LLMs), prompting them to verbalize the multimodal query into a natural-language description of the intended target for subsequent VLP-based retrieval. Although training is bypassed, the explicit query-to-text conversion inevitably discards visual details, and the generated descriptions may not conform to the textual style preferred by the VLP encoder~\cite{cliptalk}, leading to suboptimal cross-modal alignment. Moreover, such methods hinge on prompt engineering, where each task and each domain calls for carefully hand-crafted prompts, which is labor-intensive and severely undermines generalizability, precisely the property a unified framework demands.

Beyond the above textual detour, a recent line of work~\cite{magiclens,compodiff,fire,mcl} automatically synthesizes pseudo-triplets from web corpora or generative models, and uses them to pre-train a VLP-based composed retriever end-to-end. Because their pre-training objective closely mirrors the downstream task, these methods often achieve strong performance. Nevertheless, they still build upon conventional VLP-style architectures, which are optimized for single-image and short-text inputs, and are inherently incapable of modeling the temporal dynamics of video or the accumulated context of multi-turn interactions, leaving MT-CIR and CoVR out of reach.

These limitations suggest that a unified CVR framework requires a query encoder capable of directly understanding diverse multimodal inputs. MLLMs naturally satisfy this requirement, as they can process image-text, multi-turn image-text, and video-text inputs within a unified architecture. 
However, prompting them in a training-free manner would fall back into the textual detour. We thus follow the pseudo-triplet paradigm and fine-tune the MLLM into an end-to-end composed query embedder. A natural instinct is to build the retriever entirely on the MLLM, encoding both the query and the gallery with it~\cite{mmembed,vlm2vec,gme}. In the zero-shot setting, however, the pseudo-triplets are automatically synthesized and inevitably noisy, and fine-tuning both sides on such data would distort the embedding space and undermine reliable retrieval. We therefore keep the target side as a frozen, well-optimized VLP image encoder, resulting in a unified zero-shot CVR framework that leverages the complementary strengths of the two model families: the VLP model preserves its well-structured cross-modal metric space for reliable visual retrieval, while the MLLM is fine-tuned to enhance compositional and multimodal query understanding, to reason about complex visual modifications and align them effectively with the retrieval space.

However, realizing this design for strong performance across diverse downstream CVR tasks presents two key challenges. \textbf{1) Heterogeneous Embedding Spaces.} The output representations of MLLMs reside in a fundamentally different space from the well-structured metric space of VLP models. How to bridge this gap while preserving the MLLM's powerful multimodal compositional understanding on the query side and maintaining the integrity of the VLP metric space on the gallery side is the first challenge.
\textbf{2) Computational Feasibility of MLLM Reranking.} In zero-shot CVR, applying the pre-trained model to a new domain often yields suboptimal initial rankings due to domain shift. Fortunately, the initial retrieval results implicitly encode gallery distribution and candidate relevance, which can be exploited for on-the-fly ranking refinement without additional annotations. MLLMs can interpret complex visual modifications and refine the ranking accordingly~\cite{mmembed}. However, directly applying them to all candidates is computationally prohibitive. The key challenge is thus to design a mechanism that practically harnesses MLLMs' reasoning capabilities for effective reranking.

\begin{figure*}[t]
    \centering
    \includegraphics[width=\linewidth]{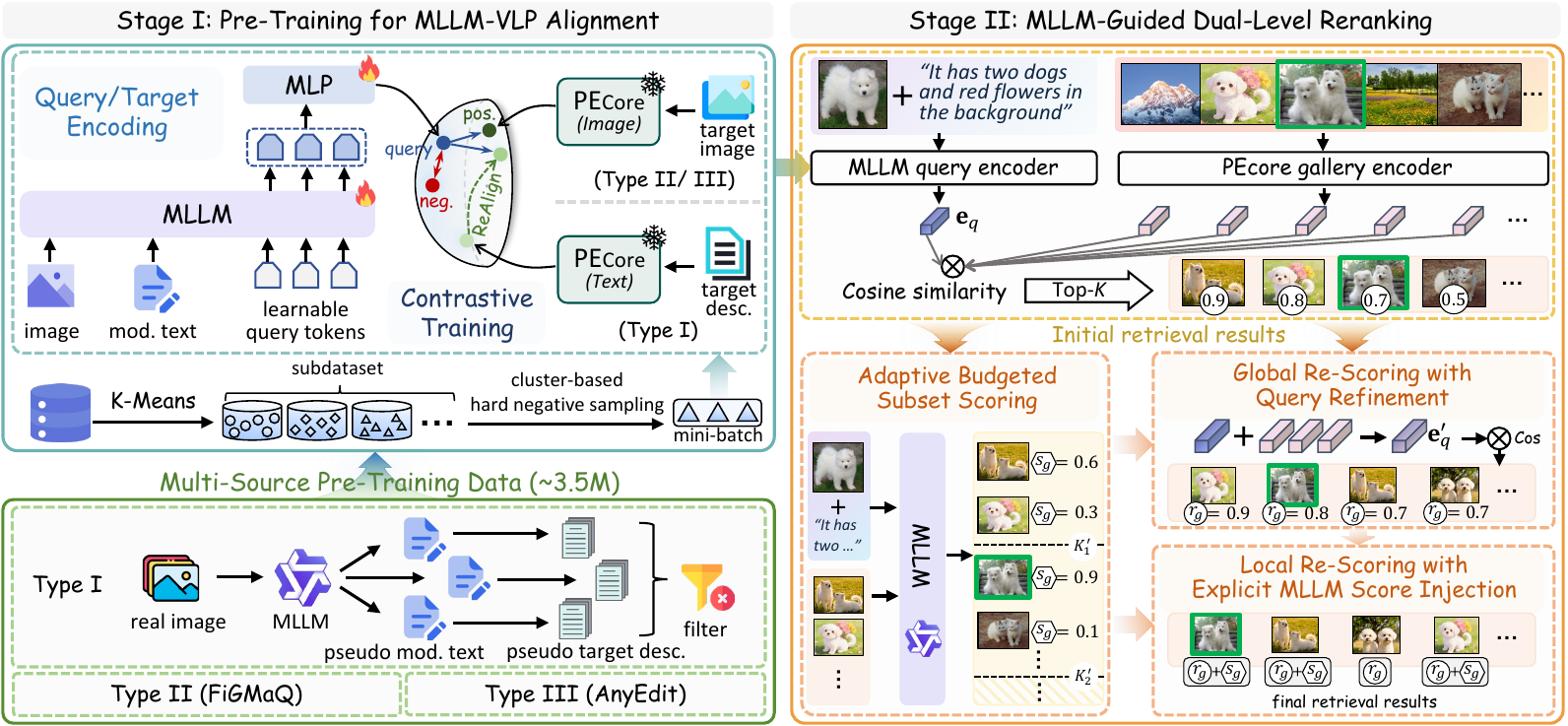}
    \caption{Overview of \method{}. \textbf{Stage~I} conducts pre-training to bridge the heterogeneous embedding spaces between the MLLM query encoder and the PEcore gallery encoder. \textbf{Stage~II} performs dual-level reranking that exploits MLLM relevance judgments over the initial retrieval results, refining the ranking through both global and local re-scoring.}
    \label{fig:framework}
\end{figure*}

In this paper, we propose \textbf{UniCVR}, a unified framework for zero-shot Composed Visual Retrieval that addresses both challenges, as shown in Figure~\ref{fig:framework}. UniCVR operates in two stages. In Stage~I, \textbf{Pre-Training for MLLM-VLP Alignment} bridges the heterogeneous embedding spaces between the MLLM query encoder and the VLP gallery encoder. An MLLM serves as the composed query encoder, natively handling image-text, multi-turn image-text, and video-text inputs, making it directly applicable to all three CVR tasks.  We specifically adopt PEcore~\cite{pecore} as the gallery encoder, which is jointly pre-trained on large-scale image-text and video-text corpora, enabling a unified gallery embedding across both images and videos. 
To provide diverse and sufficient alignment supervision, we curate a multi-source pre-training dataset of approximately 3.5M samples, with a cluster-based hard negative sampling strategy to strengthen the contrastive supervision signal. 
Notably, the dataset's effectiveness stems more from its diversity than from its scale, which is modest compared to existing pseudo-triplet datasets.
In Stage~II, \textbf{MLLM-Guided Dual-Level Reranking} leverages the reasoning capabilities of MLLMs to refine the initial retrieval results. To avoid the prohibitive cost of scoring all gallery candidates with an MLLM, we introduce an adaptive budgeted subset scoring mechanism that evaluates only a small subset of top-ranked candidates with early termination. The resulting MLLM relevance scores are then exploited through a dual-level re-scoring mechanism: a global re-scoring that refines the query embedding via score-weighted fusion of candidate embeddings and recomputes similarities for all gallery items, and a local re-scoring that directly injects the explicit MLLM relevance scores for the scored candidates, producing more accurate final rankings without incurring excessive computation.
Our contributions can be summarized as follows:
\begin{itemize}[leftmargin=*]
    \item We propose UniCVR, a unified zero-shot framework for Composed Visual Retrieval (CVR) that bridges the powerful multimodal reasoning of MLLMs with the well-structured metric space of VLP models. To the best of our knowledge, UniCVR is the first to unify CIR, MT-CIR, and CoVR within a single zero-shot framework.  
    \item We propose an MLLM-Guided Dual-Level Reranking mechanism that adaptively scores a budgeted candidate subset and exploits the resulting relevance signals through dual-level re-scoring, \ie, global query refinement and local MLLM score injection, achieving effective ranking refinement.
    \item Extensive experiments on FashionIQ, CIRR, CIRCO, Multi-Turn FashionIQ, and WebVid-CoVR validate UniCVR's effectiveness and generalizability across all three CVR tasks. Our code and dataset are released\footnote{\url{https://github.com/haokunwen/UniCVR}} to benefit the community.
\end{itemize}

\section{Related Work}

\label{sec:related}
 
\subsection{Composed Visual Retrieval}
CIR has been studied under both supervised and zero-shot settings~\cite{survey}. 
Existing ZS-CIR methods can be broadly grouped into three lines. 
\textbf{(i) Textual-inversion approaches}~\cite{pic2word,searle,isearle,fti4cir,keds,lincir} learn a lightweight mapping network that inverts the reference image into one or a few pseudo-word tokens. The pseudo-word is then concatenated with the modification text to form a purely textual query, so that the composed query can be directly fed into a text encoder of a pre-trained VLP model (\eg, CLIP~\cite{clip}) and matched against candidate images in the shared embedding space. Since the mapping network is typically optimized with image-text pairs or even images alone, these methods bypass the need for annotated triplets, yet the resulting pseudo-word is often too coarse to preserve fine-grained visual details.
\textbf{(ii) Training-free captioning approaches}~\cite{ldre,cirevl,osrcir,autocir} instead exploit the reasoning ability of (M)LLMs: the multi-modal query is first verbalized into a natural-language description of the intended target, \eg, by captioning the reference image and prompting an LLM to edit the caption according to the modification text, or by directly prompting an MLLM. Retrieval again reduces to text-to-image matching with an off-the-shelf VLP model, which requires no training but incurs non-trivial inference cost and is sensitive to the fidelity of the intermediate caption.
\textbf{(iii) Pseudo-triplet pre-training approaches}~\cite{fire,mcl,magiclens,compodiff} automatically construct large-scale triplets from web data or generative models, and pre-train a retriever end-to-end with contrastive learning. Being end-to-end, such models avoid the detour through the textual modality and are therefore more efficient at inference, yet the quality and diversity of the constructed triplets play a decisive role in their final performance.
MT-CIR~\cite{dialog,cfir,irr,fashionntm} extends CIR to iterative query refinement across multiple rounds, but all existing methods require supervised training on annotated multi-turn dialogues. CoVR~\cite{covr,covr2,covrenrich,fdca,hud} replaces the reference image with a video clip; most works focus on supervised training and ZS-CoVR remains largely unexplored. 

Despite the shared compositional query paradigm, no prior work has unified these three tasks under a single zero-shot framework. UniCVR bridges this gap by employing an MLLM as a unified compositional query encoder across all three tasks, paired with a VLP gallery encoder that supports both image and video retrieval.

\subsection{MLLMs as Rerankers}
Leveraging (M)LLMs for reranking originates from text retrieval. RankGPT~\cite{rankgpt} prompts GPT models to perform listwise passage reranking via a sliding-window strategy. Subsequent works explore pairwise prompting~\cite{pairranker}, which compares candidates two at a time to mitigate positional bias, as well as fine-tuning open-source LLMs into dedicated rerankers~\cite{rankllama}. In the multimodal domain, LamRA~\cite{lamra} adapts an MLLM as a listwise reranker for image-text retrieval by fine-tuning with ranking-oriented instructions. MM-Embed~\cite{mmembed} combines MLLM-based bi-encoder retrieval with MLLM pointwise reranking, demonstrating that the two stages are complementary. However, these methods typically apply the MLLM to a fixed candidate set without considering computational cost. 
UniCVR introduces adaptive budgeted scoring with early termination, so that easy queries are resolved with a small budget while only hard queries incur a larger one, together with a dual-level re-scoring scheme that propagates the MLLM's relevance judgments both globally and locally, making MLLM-based reranking computationally feasible for large-scale CVR galleries.
\section{Methodology}
\label{sec:method}

In this section, we first formulate the problem and then detail the UniCVR framework. Unlike existing training-free methods~\cite{ldre,autocir,seize} that use MLLMs as captioners, UniCVR treats the MLLM as a compositional query embedder, mapping the composed query directly into a dense embedding aligned with the VLP gallery space. This avoids both the information bottleneck introduced by a textual intermediary and the need for hand-designed, task-specific prompts.

\subsection{Problem Formulation}
\label{ssec:formulation}
We propose a unified framework for Zero-Shot Composed Visual Retrieval (ZS-CVR) that subsumes CIR, MT-CIR, and CoVR under a single model. In all three tasks, the composed query consists of one or more reference visuals (images or video clips), paired with a modification text describing the desired change. For single-turn CIR and CoVR, the query is a single image-text or video-text pair $\mathcal{Q} = (V_r, T_m)$; for MT-CIR, it is an interleaved sequence of $S$ reference visuals and modification texts across multiple turns, $\mathcal{Q} = (V_r^{(1)}, T_m^{(1)}, \ldots, V_r^{(S)}, T_m^{(S)})$. Given $\mathcal{Q}$ and a gallery $\mathcal{G} = \{V_g^1, V_g^2, \ldots, V_g^N\}$ of candidate visuals, the task is to retrieve the target visual $V_t \in \mathcal{G}$ that best matches the composed modification intent:
\begin{equation}
  V_t = \operatorname*{arg\,max}_{V_g \in \mathcal{G}}\;
  \phi(\mathcal{Q},\, V_g),
  \label{eq:task}
\end{equation}
where $\phi(\mathcal{Q}, V_g)$ denotes a relevance scoring function between the composed query and a gallery candidate. The \textit{zero-shot} setting requires the model to generalize to every downstream task and domain without any task-specific human-annotated triplets, with one unified model applied across all tasks.

\subsection{Stage~I: Pre-Training for MLLM-VLP Alignment}
\label{ssec:pretrain}
While the ZS-CVR setting precludes human-annotated triplets for any downstream task, readily collectible triplets can serve as a viable proxy for learning the alignment between the MLLM query encoder and the VLP gallery space. Stage~I leverages such data to train the composed query encoder via contrastive pre-training, laying the foundation for zero-shot generalization to unseen downstream tasks.

\subsubsection{Multi-Source Pre-Training Data}
\label{sssec:data}
Several existing datasets can serve as pre-training resources for zero-shot compositional retrieval, including \textbf{MMC}~\cite{mcl} and \textbf{FiGMaQ}~\cite{fire}, which have distinct data formats. MMC follows $\langle$reference image, modification text, target image caption$\rangle$, while FiGMaQ supports $\langle$reference image, modification text, target image$\rangle$. These differences offer complementary supervision for semantic transformation and target matching, respectively. Beyond these, we observe that large-scale image editing datasets~\cite{anyedit, omniedit, pix2pix} naturally satisfy the triplet structure required by ZS-CVR, and their stricter query--target alignment is particularly beneficial for the required cross-modal embedding alignment.

\begin{figure*}[ht]
\centering
\includegraphics[width=\linewidth]{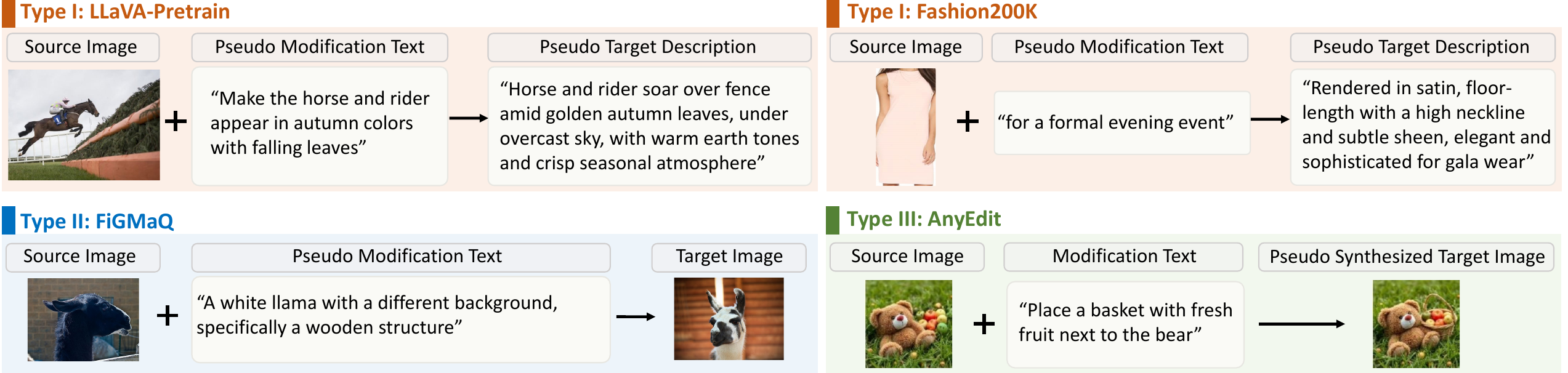}
\caption{Representative triplets from each pre-training source. \textbf{Type~I} (LLaVA-Pretrain and Fashion200K) pairs a source image with a pseudo modification text and a pseudo target description. \textbf{Type~II} (FiGMaQ) pairs a source image with a pseudo modification text and a real target image. \textbf{Type~III} (AnyEdit) pairs a source image with a modification text and a pseudo synthesized target image.}
\label{fig:data_examples}
\end{figure*}

Accordingly, we curate a multi-source dataset of approximately 3.5M samples drawn from three data types (Table~\ref{tab:data}), with one representative triplet per type shown in Figure~\ref{fig:data_examples}. These data types are complementary not only in their triplet forms, but also in the supervision they provide for MLLM--VLP alignment. Type~I supports highly diverse and open-ended modification intents, thereby broadening the semantic coverage of compositional transformations. However, its target is represented by a generated textual description rather than a real visual instance, which may introduce a modality discrepancy and generation noise. Type~II uses real image pairs as targets and therefore provides genuine visual matching supervision, but its modification intents are constrained by the naturally occurring differences between the mined image pairs and may lack sufficient diversity. Type~III provides the strongest query--target correspondence because the target image is explicitly synthesized according to the modification instruction. Nevertheless, edited targets are often visually close to their source images, making this type less effective for learning large semantic transitions or substantial changes in scene composition. These limitations are complementary: Type~I contributes semantic diversity, Type~II contributes realistic cross-instance variation, and Type~III contributes precise instruction--target correspondence. Their combination therefore provides a solid foundation for alignment pre-training.

\begin{itemize}[leftmargin=*]

\item \textbf{Type~I (with Pseudo Modification Text and Pseudo Target Description)}
follows the MMC construction pipeline rather than directly adopting MMC, since MMC relies on older models (\eg, LLaMA2~\cite{llama2}) and mainly covers the general domain, without sufficient fashion-domain data, an important application scenario for visual compositional retrieval. Specifically, we prompt Qwen3-VL-8B to generate three diverse pairs of modification text and target description per source image. The three prompts are designed to improve the diversity, visual faithfulness, and retrieval relevance of the generated transformations. We further apply lexical and semantic filtering to remove low-quality generations, including overly short or uninformative modifications, comparative expressions that depend on unavailable contextual images, and trivial transformations whose target descriptions remain excessively similar to the source image according to CLIP similarity. We use LLaVA-Pretrain~\cite{llava} as the general-domain source and incorporate Fashion200K~\cite{fashion200k} to improve fashion-domain coverage.

\item \textbf{Type~II (with Pseudo Modification Text)} directly adopts FiGMaQ~\cite{fire}, where visually correlated image pairs are mined from the ImageNet1K test split~\cite{imagenet}, and modification texts are generated by LLaMA 3.1-70B. 

\item \textbf{Type~III (with Pseudo Synthesized Target Image)} leverages AnyEdit~\cite{anyedit}, a large-scale image editing dataset with high-quality edited images. It naturally forms triplets $\langle$real image, modification text, edited image$\rangle$, and we retain only retrieval-relevant categories (object replacement, color alteration, appearance swap), discarding less relevant operations such as depth modification.
\end{itemize}

\begin{table}[!ht]
\centering
\caption{Overview of the multi-source pre-training dataset.}
\label{tab:data}
\fit{
\begin{tabular}{llccc}
\toprule
Type & Source & Triplet Form & Domain & \#Samples \\
\midrule
Type~I   & LLaVA-Pretrain~\cite{llava}       & $\langle$real img, mod. text, target desc.$\rangle$    & General & 1,430K \\
Type~I   & Fashion200K~\cite{fashion200k}    & $\langle$real img, mod. text, target desc.$\rangle$    & Fashion & 515K   \\
Type~II  & FiGMaQ~\cite{fire}                & $\langle$real img, mod. text, real target img$\rangle$ & General & 87K    \\
Type~III & AnyEdit~\cite{anyedit} (filtered) & $\langle$real img, mod. text, edited img$\rangle$      & General & 1,496K \\
\midrule
Total    &                                   &                                                        &         & $\sim$3.5M \\
\bottomrule
\end{tabular}
}
\end{table}

\subsubsection{Query/Target Encoding}
\label{sssec:encoding}

\noindent\textbf{MLLM-based Query Encoding.}
We adopt Qwen3-VL~\cite{qwen3vl}, the powerful open-source MLLM, as the composed query encoder. The reference visual and the modification text are fed to the MLLM together with a task instruction prompt that guides the model to reason about the intended target. To extract a compact retrieval-oriented representation, we append $M$ learnable query tokens $\{\mathbf{q}_1, \ldots, \mathbf{q}_M\}$ to the end of the input sequence~\cite{metaqueries, fire}. By virtue of causal self-attention, these tokens attend over all preceding context and holistically aggregate the composed query semantics. The output hidden states at the $M$ query token positions are mean-pooled and projected into the gallery embedding space via a lightweight MLP. This can be formulated as follows,
\begin{equation}
  \begin{cases}
    \{\mathbf{h}_i\}_{i=1}^{M} = \mathrm{MLLM}(\mathcal{Q},\ \mathbf{q}_1, \ldots, \mathbf{q}_M), \\[6pt]
    \mathbf{e}_q = \mathrm{MLP}\!\left(\dfrac{1}{M}\sum_{i=1}^{M}\mathbf{h}_i\right),
  \end{cases}
  \label{eq:query_embed}
\end{equation}
where $\mathbf{e}_q \in \mathbb{R}^D$ is the query embedding. 

\noindent\textbf{VLP-based Target Encoding.} 
For the target encoder, we adopt PEcore~\cite{pecore}, an advanced VLP model jointly pre-trained on large-scale image-text and video-text corpora. Unlike CLIP, which is trained solely on image-text pairs, PEcore provides a unified embedding space for both still images and video clips, enabling a single target index across CIR, MT-CIR, and CoVR without separate models per task. 

For Types~II and~III, whose target is a visual $V_t$ (an edited or real image), we directly use the frozen PEcore image encoder to derive the target embedding as follows,
\begin{equation}
  \mathbf{e}_t = \mathrm{PEcore}(V_t),
  \label{eq:gallery_embed}
\end{equation}
where $\mathbf{e}_t \in \mathbb{R}^{D}$ is the target embedding, sharing the same dimensionality with $\mathbf{e}_q$.

While for Type~I, since the retrieval target is a textual description $D_t$, directly using $\mathrm{PEcore}_\mathrm{txt}(D_t)$ as the retrieval target would cause a training-test discrepancy, as text and image embeddings reside on different manifolds even in well-trained VLP models~\cite{realign,gap}. Here we adopt the ReAlign transformation~\cite{realign} to bridge this gap, which shifts the text embedding onto the image manifold via a pre-computed affine mapping followed by $\ell_2$ normalization:
\begin{equation}
  \begin{cases}
    \tilde{\mathbf{e}}_t = (\mathrm{PEcore}_\mathrm{txt}(D_t) -
    \boldsymbol{\mu}_\mathrm{txt}) \cdot \sqrt{\dfrac{\mathrm{tr}(\boldsymbol{\Sigma}_\mathrm{img})}{\mathrm{tr}(\boldsymbol{\Sigma}_\mathrm{txt})}} + \boldsymbol{\mu}_\mathrm{img}, \\[8pt]
    {\mathbf{e}}_t = \ell_2(\tilde{\mathbf{e}}_t),
  \end{cases}
  \label{eq:realign}
\end{equation}
where $\boldsymbol{\mu}_\mathrm{txt}$ and $\boldsymbol{\mu}_\mathrm{img}$ are the mean text and image embeddings, $\boldsymbol{\Sigma}_\mathrm{txt}$ and $\boldsymbol{\Sigma}_\mathrm{img}$ are their covariance matrices, $\mathrm{tr}(\cdot)$ denotes the matrix trace, and $\ell_2(\cdot)$ denotes $\ell_2$ normalization. The scaling factor $\sqrt{\mathrm{tr}(\boldsymbol{\Sigma}_\mathrm{img})/\mathrm{tr}(\boldsymbol{\Sigma}_\mathrm{txt})}$ normalizes the variance difference between the two modalities. All statistics are pre-computed offline from $10{,}000$ randomly sampled texts and images and fixed during training, so the transformation introduces no additional training cost. ${\mathbf{e}}_t \in \mathbb{R}^D$ is the resulting target embedding aligned to the image manifold, and we empirically verify that it does close the modality gap in Section~\ref{sec:data-analysis}.

\subsubsection{Contrastive Training}
\label{sssec:contrastive}
We adopt contrastive training to optimize the alignment between the MLLM-based query encoder and the PEcore target space. Since the contrastive loss relies on in-batch negatives for supervision, the quality of negatives is critical. With random sampling across our heterogeneous multi-source data, negatives are likely to be visually distant from the positive target, yielding weak gradient signal. To address this, we propose a \textit{cluster-based hard negative sampling strategy}: we first partition each source dataset into data clusters based on target embedding similarity, then constrain each mini-batch to samples from a single cluster, ensuring hard contrastive supervision throughout training.

Specifically, before training, we apply the $K$-Means algorithm~\cite{kmeans} to each source dataset using the pre-computed target embeddings $\mathbf{e}_t$, ensuring semantically similar samples are grouped within the same cluster. All cluster assignments and target embeddings are cached offline. During training, each mini-batch consists of triplets sampled from a single cluster $\mathcal{D}_{k(t)}$. We denote the $i$-th query-target pair as $\mathbf{e}_q^i, \mathbf{e}_t^i \in \mathcal{D}_{k(t)}$, where $k(t)$ is the cluster index assigned to training step $t$. The InfoNCE loss over the mini-batch is:
\begin{equation}
\mathcal{L}_{k(t)} = -\frac{1}{B}\sum_{i=1}^{B}
  \log \frac{\exp(\mathrm{sim}(\mathbf{e}_q^i, \mathbf{e}_t^i)/\tau)}
           {\sum_{j=1}^{B} \exp(\mathrm{sim}(\mathbf{e}_q^i, \mathbf{e}_t^j)/\tau)},
\end{equation}
where $\mathrm{sim}(\cdot, \cdot)$ denotes cosine similarity and $\tau$ is a learnable temperature parameter. Note that the optimized parameters comprise only the learnable query tokens, the lightweight MLP, and the LoRA~\cite{lora} matrices inserted into the MLLM, while PEcore remains frozen.
Freezing the target encoder is a deliberate choice, motivated by two considerations. First, the triplets available in the zero-shot setting are automatically curated and inevitably noisy; making the target encoder trainable would let this noise reshape the target space, eroding the very metric structure that VLP pre-training provides and that retrieval relies on. Second, a frozen target encoder allows all gallery embeddings to be pre-computed offline, so that each training step back-propagates through the query branch alone, which is what makes scaling the query encoder feasible under our compute budget.

Upon completion of Stage~I, the query encoder and PEcore gallery encoder together enable
zero-shot retrieval across all downstream tasks, producing the initial ranking and the query embedding that Stage~II consumes.

\subsection{Stage~II: MLLM-Guided Dual-Level Reranking}
\label{ssec:rerank}
To enhance zero-shot CVR performance at test time, we propose an MLLM-guided dual-level reranking strategy. Unlike prior works that rely on shallow networks to adjust initial rankings, our approach leverages the reasoning capabilities of MLLMs to refine retrieval results. While several studies have highlighted the potential of MLLMs as powerful rerankers~\cite{lamra,mmembed}, they often overlook the computational feasibility: applying an MLLM to a large-scale gallery is computationally prohibitive. To address this, we first perform adaptive budgeted subset scoring, estimating relevance only for a subset of top-ranked candidates using the MLLM. We then employ a dual-level mechanism for adjusting candidate relevance scores: a global adjustment that refines the query embedding via a score-weighted fusion of the MLLM-assessed candidate embeddings, and a local adjustment that directly injects the explicit MLLM relevance scores. This design balances computational cost and accuracy, enabling practical and reliable reranking.

\subsubsection{Adaptive Budgeted Subset Scoring} 
To reduce retrieval-specific bias from Stage~I and obtain more objective relevance signals, we use an MLLM $\mathcal{A}$ with LoRA adapters disabled as the relevance assessor. Specifically, given a query $\mathcal{Q}$ and a gallery image $V_g\in \mathcal{G}$, the MLLM is prompted with a binary Yes/No instruction $P_s$:
\textit{``Does the candidate match the reference modified by the given text? Answer Yes or No.''} The relevance score for each candidate is computed as:
\begin{equation}
  \begin{cases}
    z_{\text{``Yes''}}, z_{\text{``No''}} = \mathcal{A}(V_r, T_m, V_g, P_s),\\[2mm]
    s_g = \sigma(z_{\text{``Yes''}} - z_{\text{``No''}}),
  \end{cases}
  \label{eq:score}
\end{equation}
where $z_{\text{``Yes''}}$ and $z_{\text{``No''}}$ are the output logits for the Yes/No tokens, $\sigma(\cdot)$ is the sigmoid function, and $s_g \in (0,1)$ is the resulting continuous relevance score.

Apparently, directly scoring all $K$ candidates using the MLLM is computationally prohibitive. To address this, we adopt an adaptive budgeted subset scoring strategy, which leverages an MLLM to independently evaluate relevance for a subset of top-ranked candidates. The intuition is that top-ranked candidates are more likely to be relevant, and their embeddings provide higher-quality signals for refining the query representation, enabling more accurate global matching assessment across the gallery.

Specifically, we define two candidate subset sizes, $K'_1 < K'_2 \leq K$, and initially score only the top-$K'_1$ candidates. If at least one candidate achieves a score above a confidence threshold $\delta$, we consider that a sufficiently relevant candidate has been identified and can be reliably used to refine the query representation. The scoring process can then be terminated early, avoiding additional computational cost. Otherwise, scoring is extended to the top-$K'_2$ candidates. Formally, we denote the adaptively determined scoring budget as $K'$, i.e., $K' = K'_1$ if early termination occurs, and $K' = K'_2$ otherwise. $\mathcal{C}_{K'}$ denotes the subset of top-ranked candidates whose relevance has been independently assessed by the MLLM, used in the subsequent dual-level candidate relevance adjustment procedure.  

\subsubsection{Global Re-Scoring with Query Refinement}
Since we only obtain MLLM scores for a subset of candidates, direct score-level fusion across the entire gallery, i.e., combining the original Stage I scores with Stage II MLLM scores, is not feasible. To address this limitation, we propose a global re-scoring strategy based on intermediate query refinement, that utilizes the MLLM objectively-assessed relevance signals to refine the query embedding. The refined embedding is then used to recompute relevance scores for all candidates in the gallery, effectively propagating the refined signal globally.

Intuitively,  candidates with higher scores output by the assessor $\mathcal{A}$ are deemed more reliably relevant to the query, and thus contribute more to the updated query representation. Accordingly, we refine the query embedding by computing a score-weighted centroid of the MLLM-scored $K'$ candidate embeddings and interpolating it with the original query embedding, pulling the query embedding toward the visual neighborhood of the most relevant candidates, enabling more accurate re-scoring of all gallery items:
\begin{equation}
  \mathbf{e}_q^{\prime} = \ell_2\!\left((1-\alpha)\,\mathbf{e}_q +
  \alpha \cdot \frac{\sum_{V_g\in \mathcal{C}_{K'}} s_g\, \mathbf{e}_g}{\sum_{V_g\in \mathcal{C}_{K'}} s_g}\right),
  \label{eq:expansion}
\end{equation}
where $\mathbf{e}_g = \mathrm{PEcore}(V_g)$ is the pre-computed gallery embedding of $V_g$, $\ell_2(\cdot)$ denotes $\ell_2$ normalization, and $\alpha \in [0,1]$ is an interpolation coefficient. $\mathbf{e}_q^{\prime}$ is the refined query embedding, used as follows for rescoring the global gallery images:
\begin{equation}
  r_g= \mathrm{sim}(\mathbf{e}_q^{\prime}, \mathbf{e}_g),
\end{equation}
where $r_g$ is the refined score of each gallery image $V_g\in \mathcal{G}$. 

\noindent\textbf{}

\noindent\subsubsection{Local Re-Scoring with Explicit MLLM Score Injection}
While global rescoring using the refined query embedding enhances similarity scores across the gallery, it does not directly leverage the explicit relevance judgments provided by the MLLM for candidates in $\mathcal{C}_{K'}$. These MLLM-derived scores offer a more direct and reliable measure of query-candidate correspondence than embedding similarity alone.
We therefore fuse the two signals within the scored window via  a weighted interpolation:
\begin{equation}   
    r_g^*= \begin{cases}
    (1-\beta)\,\dfrac{r_g - r_{\min}}
                      {r_{\max} - r_{\min}} + \beta\, s_g,   & \text{if } V_g \in \mathcal{C}_{K'}, \\[10pt]
    {r}_g, & \text{if }  V_g\in \mathcal{G} \setminus \mathcal{C}_{K'},
  \end{cases}
  \label{eq:rerank}
\end{equation}
where $r_{\min}$ and $r_{\max}$ are the minimum and maximum cosine similarities within $\mathcal{C}_{K'}$. Min-max normalization projects $r_g$ to the same scale as $s_g \in (0, 1)$. $\beta \in [0,1]$ is an interpolation coefficient. Candidates outside the scored subset retain their recomputed cosine similarities unchanged. The gallery is then re-ranked by $r_g^*$ to produce the final retrieval results.

\section{Experiment}
\label{sec:exp}
In this section, we first detail the experimental settings, followed by a comprehensive analysis of the results.

\begin{table*}[!ht]
    \centering
    \caption{Performance comparison on the validation set of FashionIQ. ``Par.''\ denotes the zero-shot paradigm: TI = textual inversion, PT = pseudo-triplet pre-training, TF = training-free. The best results are in bold, while the second-best results are underlined.}
    \label{tab:exp_fashioniq}
    \fit{
    \begin{tabular}{l|c|c|c|cc|cc|cc|cc|c}
    \hline
    \multirow{2}{*}{Model} & \multirow{2}{*}{Par.} & \multirow{2}{*}{\makecell{Query \\ Reasoner}} & \multirow{2}{*}{Retriever} & \multicolumn{2}{c|}{Dresses} & \multicolumn{2}{c|}{Shirts} & \multicolumn{2}{c|} {Tops\&Tees} & \multicolumn{2}{c|}{Average} & \multirow{2}{*}{Avg.} \\ \cline{5-12}
    & & & & R@$10$ & R@$50$ & R@$10$ & R@$50$ & R@$10$ & R@$50$ & R@$10$ & R@$50$ & \\
    \hline \hline
    SEARLE-XL~\cite{searle} \footnotesize{\textcolor{gray}{(ICCV'23)}} & TI & CLIP-L (T) & CLIP-L & $20.48$ & $43.13$ & $26.89$ & $45.58$ & $29.32$ & $49.97$ & $25.56$ & $46.23$ & $ 35.90 $ \\
    iSEARLE-XL~\cite{isearle} \footnotesize{\textcolor{gray}{(TPAMI'25)}} & TI & CLIP-L (T) & CLIP-L & $22.51$ & $46.36$ & $28.75$ & $47.84$ & $31.31$ & $52.68$ & $27.52$ & $48.96$ & $ 38.24 $\\
    FTI4CIR~\cite{fti4cir} \footnotesize{\textcolor{gray}{(SIGIR'24)}} & TI & CLIP-L (T) & CLIP-L & $24.39$ & $47.84$ & $31.35$ & $50.59$ & $32.43$ & $54.21$ & $29.39$ & $50.88$ & $40.14$\\
    KEDs~\cite{keds} \footnotesize{\textcolor{gray}{(CVPR'24)}} & TI & CLIP-L (T) & CLIP-L & $21.70$ & $43.80$ & $28.90$ & $48.00$ & $29.90$ & $51.90$ & $26.80$ & $47.90$ & $37.37$ \\
    MagicLens~\cite{magiclens} \footnotesize{\textcolor{gray}{(ICML'24)}} & PT & CoCa-L & CoCa-L & $32.30$ & $52.70$ & $40.50$ & $59.20$ & $41.40$ & $63.00$ & $38.00$ & $58.20$ & $48.10$\\
    CompoDiff~\cite{compodiff} \footnotesize{\textcolor{gray}{(TMLR'24)}} & PT & DN-Transformer & CLIP-G & \underline{$37.78$} & $49.10$ & $41.31$ & $55.17$ & $44.26$ & $56.41$ & $41.12$ & $53.56$ & $47.34$\\
    LDRE~\cite{ldre} \footnotesize{\textcolor{gray}{(SIGIR'24)}}& TF & GPT-3.5-Turbo & CLIP-G & $26.11$ & $51.12$ & $35.94$ & $58.58$ & $35.42$ & $56.67$ & $32.49$ & $55.46$ & $43.97$ \\
    CIReVL~\cite{cirevl} \footnotesize{\textcolor{gray}{(ICLR'24)}} & TF & GPT-3.5-Turbo &  CLIP-G  & $ 27.07 $ & $ 49.53 $ & $ 33.71 $ & $ 51.42 $ & $ 35.80 $ & $ 56.14 $ & $ 32.19 $ & $ 52.36 $ & $ 42.28 $ \\
    OSrCIR~\cite{osrcir} \footnotesize{\textcolor{gray}{(CVPR'25)}} & TF & GPT-3.5-Turbo & CLIP-G     & $33.02$ & $54.78$ & $38.65$ & $54.71$ & $41.04$ & $61.83$ & $37.57$ & $57.11$ & $47.34$ \\
    FiRE~\cite{fire} \footnotesize{\textcolor{gray}{(SIGIR'25)}} & PT & BLIP-3 & BLIP-3     & $29.60$ & $50.87$ & $39.84$ & $60.06$ & $35.64$ & $57.83$ & $35.02$ & $56.25$ & $ 45.64 $ \\
    MOA~\cite{moa}  \footnotesize{\textcolor{gray}{(SIGIR'25)}} & PT & BLIP & BLIP     & $26.80$ & $49.10$ & $33.10$ & $52.40$ & $35.70$ & $57.10$ & $31.90$ & $52.80$ & $ 42.35 $ \\
    CoTMR~\cite{cotmr} \footnotesize{\textcolor{gray}{(ICCV'25)}} & TF & Qwen2-VL-72B  &  CLIP-G  & $ 34.51 $ & $ 57.36 $ & $ 38.32 $ & $ 62.24 $ & $ 41.90 $ & $ 64.30 $ & $ 38.25 $ & $ 61.32 $ & $ 49.79 $ \\
    AutoCIR~\cite{autocir} \footnotesize{\textcolor{gray}{(KDD'25)}} & TF & GPT-4o-Mini  &  CLIP-G  & $26.18$ & $ 47.69 $ & $ 36.36 $ & $ 55.84 $ & $ 37.28 $ & $ 60.38 $ & $ 33.27 $ & $ 54.63 $ & $ 43.95 $ \\
    G-MIXER~\cite{gmixer} \footnotesize{\textcolor{gray}{(CVPR'26)}} & TF & GPT-4o  &  CLIP-G  & $34.71$ & $ 58.85 $ & $ 39.65 $ & $ 59.61 $ & $ 44.77 $ & $ 67.47 $ & $ 39.71 $ & $ 61.98 $ & $ 50.85 $ \\
    \hline
    \rowcolor{blue!8}
    UniCVR (Stage~I)  & PT & Qwen3-VL-32B   & PEcore-G     & $37.18$ & \underline{$60.04$} & \underline{$43.62$} & \underline{$63.74$} & \underline{$46.35$} & \underline{$67.77$} & \underline{$42.38$} & \underline{$63.85$} & \underline{$53.12$} \\
    \rowcolor{blue!8}
    UniCVR (Stage~II) & PT & Qwen3-VL-32B & PEcore-G     & \textbf{42.04} & \textbf{60.88} & \textbf{46.27} & \textbf{63.84} & \textbf{49.72} & \textbf{68.23} & \textbf{46.01} & \textbf{64.32} & \textbf{55.17} \\
    \hline
    \end{tabular}
    }
\end{table*}

\begin{table*}[t]
    \centering
    \caption{Performance comparison on CIRR and CIRCO. ``Par.''\ follows the legend of Table~\ref{tab:exp_fashioniq}.}
    \label{tab:exp_cirr_circo}
    \fit{
    \begin{tabular}{l|c|c|c|cccc|ccc|cccc}
    \hline
    \multirow{3}{*}{Model} & \multirow{3}{*}{Par.} & \multirow{3}{*}{\makecell{Query \\ Reasoner}} & \multirow{3}{*}{Retriever} & \multicolumn{7}{c|}{CIRR} & \multicolumn{4}{c}{CIRCO}  \\
    \cline{5-15}
    &&& &\multicolumn{4}{c|}{{R@$k$}} & \multicolumn{3}{c|}{{R\textsubscript{subset}@$k$}} & \multicolumn{4}{c}{{mAP@$k$}}\\
    \cline{5-15}
    & & & & $k=1$ & $k=5$ & $k=10$ & $k=50$ & $k=1$ & $k=2$ & $k=3$ & $k=5$ & $k=10$ & $k=25$ & $k=50$ \\
    \hline \hline
    SEARLE-XL~\cite{searle} \footnotesize{\textcolor{gray}{(ICCV'23)}} & TI & CLIP-L (T) & CLIP-L & $24.24 $ & $52.48$ & $66.29$ & $88.84$ & $53.76$ & $75.01$ & $88.19$ & $11.68$ & $12.73$ & $14.33$ &$15.12$\\
    iSEARLE-XL~\cite{isearle} \footnotesize{\textcolor{gray}{(TPAMI'25)}} & TI & CLIP-L (T) & CLIP-L & $25.28$ & $54.00$ & $66.72$ & $88.80$ & $-$ & $-$ & $-$ & $12.50$ & $13.61$ & $15.36$ & $16.25$\\
    FTI4CIR~\cite{fti4cir} \footnotesize{\textcolor{gray}{(SIGIR'24)}} & TI & CLIP-L (T) & CLIP-L & $25.90$ & $55.61$ & $67.66$ & $89.66$ & $55.21$ & $75.88$ & $87.98$ & $15.05$ & $16.32$ & $18.06$ & $19.05$\\
    KEDs~\cite{keds} \footnotesize{\textcolor{gray}{(CVPR'24)}} & TI & CLIP-L (T) & CLIP-L & $26.40$ & $54.80$ & $67.20$ & $89.20$ & $-$ & $-$ & $-$ & $-$ & $-$ & $-$ & $-$\\
    MagicLens~\cite{magiclens} \footnotesize{\textcolor{gray}{(ICML'24)}} & PT & CoCa-L & CoCa-L & $33.30$ & $67.00$ & $77.90$ & $94.40$ & $70.90$ & $87.30$ & $94.50$ & $34.10$ & $35.40$ & $38.10$ & $39.20$\\
    MCL~\cite{mcl} \footnotesize{\textcolor{gray}{(ICML'24)}} & PT & Llama2-7B & CLIP-L & $26.22$ & $56.84$ & $-$ & $91.35$ & $61.45$ & $-$ & $-$ & $17.67$ & $18.86$ & $20.80$ & $21.68$\\
    CompoDiff~\cite{compodiff} \footnotesize{\textcolor{gray}{(TMLR'24)}} & PT & DN-Transformer & CLIP-G & $26.71$ & $55.14$ & $74.52$ & $92.01$ & $64.54$ & $82.39$ & $91.81$ & $15.33$ & $17.71$ & $19.45$ & $21.01$\\
    LDRE~\cite{ldre} \footnotesize{\textcolor{gray}{(SIGIR'24)}}& TF & GPT-3.5-Turbo & CLIP-G & $36.15$ & $66.39$ & $77.25$ & $93.95$ & $68.82$ & $85.66$ & $93.76$ & $31.12$ & $32.24$ & $34.95$ & $36.03$\\
    CIReVL~\cite{cirevl} \footnotesize{\textcolor{gray}{(ICLR'24)}} & TF & GPT-3.5-Turbo &  CLIP-G & $34.65$ & $64.29$ & $75.06$ & $91.66$ & $67.95$ & $84.87$ & $93.21$ & $26.77$ & $27.59$ & $29.96$ & $31.03$\\
    OSrCIR~\cite{osrcir} \footnotesize{\textcolor{gray}{(CVPR'25)}} & TF & GPT-3.5-Turbo & CLIP-G  & $37.26$ & $67.25$ & $77.33$ & $85.28$ & $69.22$ & $85.28$ & $93.55$  & $30.47$ & $31.14$ & $35.03$ & $36.59$ \\
    FiRE~\cite{fire} \footnotesize{\textcolor{gray}{(SIGIR'25)}} & PT & BLIP-3 & BLIP-3 & \underline{$43.33$} & \underline{$74.02$} & \underline{$83.51$} & \textbf{95.83} & \underline{$73.01$} & \textbf{88.38} & \textbf{94.94}  & $31.03$ & $32.08$ & $34.40$ & $35.50$ \\
    MOA~\cite{moa}  \footnotesize{\textcolor{gray}{(SIGIR'25)}} & PT & BLIP & BLIP  & $34.10$ & $64.90$ & $76.30$ & $93.50$ & - & - & -  & $16.10$ & $17.40$ & $19.70$ & $20.50$ \\
    CoTMR~\cite{cotmr} \footnotesize{\textcolor{gray}{(ICCV'25)}} & TF & Qwen2-VL-72B  &  CLIP-G  & $36.36$ & $67.52$ & $77.82$ & $93.99$ & $71.19$ & $86.34$ & $93.87$ & $32.23$ & $32.72$ & $35.60$ & $36.83$ \\
    AutoCIR~\cite{autocir} \footnotesize{\textcolor{gray}{(KDD'25)}} & TF & GPT-4o-Mini  &  CLIP-L  & $31.81$ & $61.95$ & $73.86$ & $92.07$ & $67.21$ & $84.89$ & $93.13$ & $24.05$ & $25.14$ & $27.35$ & $28.36$ \\
    G-MIXER~\cite{gmixer} \footnotesize{\textcolor{gray}{(CVPR'26)}} & TF & GPT-4o  &  CLIP-G  & $39.18$ & $69.83$ & $79.35$ & - & $72.36$ & $87.25$ & $93.85$ & $31.79$ & $32.54$ & $35.49$ & $36.87$ \\
    \hline
    \rowcolor{blue!8}
    UniCVR (Stage~I)  & PT & Qwen3-VL-32B   & PEcore-G     & $40.10$ & $71.16$ & $81.76$ & $95.37$ & $68.53$ & $86.00$ & $93.61$ & \underline{$34.97$} & \underline{$36.40$} & \underline{$39.11$} & \underline{$40.29$} \\
    \rowcolor{blue!8}
    UniCVR (Stage~II) & PT & Qwen3-VL-32B & PEcore-G     & \textbf{47.93} & \textbf{76.19} & \textbf{84.70} & \underline{$95.59$} & \textbf{73.83} & \underline{$88.05$} & \underline{$94.36$} & \textbf{43.09} & \textbf{43.86} & \textbf{46.24} & \textbf{47.31} \\
    \hline
    \end{tabular}
    }
\end{table*}

\subsection{Experimental Setup}
\label{ssec:setup}

\subsubsection{Datasets and Evaluation Metrics}
We evaluate UniCVR on five benchmarks spanning all three CVR tasks. For ZS-CIR, we adopt \textbf{FashionIQ}~\cite{fashioniq}, a fashion-domain CIR dataset covering three clothing categories (Dresses, Shirts, and Tops\&Tees); \textbf{CIRR}~\cite{cirr}, a general-domain CIR dataset; and \textbf{CIRCO}~\cite{searle}, a CIR benchmark with multiple ground truths per query. Following~\cite{fti4cir,searle}, we report R@10 and R@50 on the FashionIQ validation set, R@$k$ ($k\in\{1,5,10,50\}$) and R\textsubscript{subset}@$k$ ($k\in\{1,2,3\}$) on the CIRR test set, and mAP@$k$ ($k\in\{5,10,25,50\}$) on the CIRCO test set. For MT-CIR, where no prior zero-shot attempts exist, we adopt \textbf{Multi-Turn FashionIQ}~\cite{cfir}, which extends FashionIQ to the multi-turn setting, and report R@5 and R@8 across three clothing categories following supervised works~\cite{cfir,irr,fashionntm}. For CoVR, we adopt the mainstream benchmark \textbf{WebVid-CoVR}~\cite{covr} and report R@$k$ ($k\in\{1,5,10,50\}$) following existing works~\cite{covr,covr2}.

\subsubsection{Implementation Details}
\label{sssec:impl}
For the model architecture, we adopt Qwen3-VL-32B~\cite{qwen3vl} as the MLLM query encoder, with LoRA fine-tuning parameters set to $r=8$ and $\alpha=16$. Inspired by the experimental analysis in~\cite{metaqueries}, we set the number of learnable query tokens as $M=128$. The frozen gallery encoder is implemented via PEcore-G~\cite{pecore}, which generates $1280$-dimensional embeddings.
For Stage~I, we train on our curated multi-source dataset of approximately 3.5M samples using the AdamW optimizer~\cite{adamw} with a learning rate of $1\times10^{-5}$ for the LoRA parameters and $1\times10^{-4}$ for the remaining parameters, along with a cosine annealing schedule for 1 epoch. We adopt GradCache~\cite{gradcache} to expand the effective mini-batch size to $512$. For cluster-based hard negative sampling, we apply $K$-Means clustering with $K=64$ clusters for LLaVA-Pretrain (Type~I), Fashion200K (Type~I), and AnyEdit (Type~III), and $K=32$ for FiGMaQ (Type~II). The contrastive temperature $\tau$ is initialized to $0.05$ and learned during training. The pre-training is conducted on 8 NVIDIA RTX PRO 6000 GPUs.
For Stage~II, we set the adaptive scoring budgets to $K'_1=10$ and $K'_2=20$ for Multi-Turn FashionIQ, where evaluation only considers R@5 and R@8, and to $K'_1=20$ and $K'_2=40$ for all other benchmarks. The confidence threshold is set to $\delta=0.7$. The interpolation coefficients for dual-level re-scoring are empirically set to $\alpha=0.2$ and $\beta=0.3$.

\subsection{Performance Comparison}
\label{ssec:comparison}
Since no prior zero-shot work unifies all three CVR tasks, no single baseline set applies across the board. We therefore select baselines per task, prioritizing zero-shot methods and including supervised ones for reference where zero-shot competitors are scarce. The baselines are organized as follows.

\begin{itemize}[leftmargin=*]
\item \textbf{ZS-CIR (FashionIQ, CIRR, CIRCO).}
This is the most extensively studied setting, so all baselines are zero-shot, spanning the three paradigms. The \emph{textual-inversion} methods include SEARLE-XL~\cite{searle}, iSEARLE-XL~\cite{isearle}, FTI4CIR~\cite{fti4cir}, and KEDs~\cite{keds}. The \emph{pseudo-triplet pre-training} methods include MagicLens~\cite{magiclens}, CompoDiff~\cite{compodiff}, MCL~\cite{mcl}, FiRE~\cite{fire}, and MOA~\cite{moa}. The \emph{training-free} methods include LDRE~\cite{ldre}, CIReVL~\cite{cirevl}, OSrCIR~\cite{osrcir}, CoTMR~\cite{cotmr}, AutoCIR~\cite{autocir}, and G-MIXER~\cite{gmixer}.

\item \textbf{MT-CIR (Multi-Turn FashionIQ).}
No zero-shot method exists for this task, so we report supervised baselines for reference, including Dialog Manager~\cite{dialog}, CFIR~\cite{cfir}, IRR~\cite{irr}, FashionNTM~\cite{fashionntm}, and MAI~\cite{mai}. For a genuine zero-shot comparison point, we additionally reproduce ImageScope~\cite{imagescope}, a training-free method that orchestrates several (M)LLMs for diverse image retrieval tasks, including a conversational setting close to our multi-turn queries.

\item \textbf{CoVR (WebVid-CoVR).}
We report supervised baselines for reference (CoVR~\cite{covr}, CoVR-Enrich~\cite{covrenrich}, CoVR-2~\cite{covr2}, FDCA~\cite{fdca}, and HUD~\cite{hud}), alongside the available zero-shot variants of CoVR~\cite{covr} and CoVR-2~\cite{covr2}, as well as the dedicated method MoRe~\cite{more}.
\end{itemize}

Reproduced results are marked with $\dagger$ in the corresponding tables. Since methods often come in several backbone variants with differing performance, we report each baseline in its best-performing configuration. The results are summarized in Tables~\ref{tab:exp_fashioniq} and~\ref{tab:exp_cirr_circo} (CIR), Table~\ref{tab:exp_mt-fashioniq} (MT-CIR), and Table~\ref{tab:exp_web} (CoVR). We highlight the following observations.

\begin{table*}[ht]
    \centering
    \caption{Performance comparison on Multi-Turn FashionIQ. Methods marked with $\dagger$ are our own reproduced results.}
    \label{tab:exp_mt-fashioniq}
    \fit{
    \begin{tabular}{l|c|cc|cc|cc|cc|c}
    \hline
    \multirow{2}{*}{Model} & \multirow{2}{*}{\makecell{Zero-Shot}} & \multicolumn{2}{c|}{Dresses} & \multicolumn{2}{c|}{Shirts} & \multicolumn{2}{c|} {Tops\&Tees} & \multicolumn{2}{c|}{Average} & \multirow{2}{*}{Avg.} \\ \cline{3-10}
    & & R@$5$ & R@$8$ & R@$5$ & R@$8$ & R@$5$ & R@$8$ & R@$5$ & R@$8$ & \\
    \hline \hline
    Dialog Manager~\cite{dialog} \footnotesize{\textcolor{gray}{(NeurIPS'18)}} & \ding{55} & $12.7$ & $16.7$ & $13.9$ & $17.7$ & $11.6$ & $15.8$ & $12.7$ & $16.7$ & $14.7$  \\
    CFIR~\cite{cfir} \footnotesize{\textcolor{gray}{(SIGIR'21)}} & \ding{55} & $29.8$ & $33.5$ & $30.5$ & $34.1$ & $29.4$ & $33.6$ & $29.9$ & $33.7$ & $31.8$  \\
    IRR~\cite{irr} \footnotesize{\textcolor{gray}{(MM'23)}} & \ding{55} & $26.8$ & $31.2$ & $25.8$ & $30.4$ & $27.1$ & $31.7$ & $26.6$ & $31.1$ & $ 28.8 $  \\
    FashionNTM~\cite{fashionntm} \footnotesize{\textcolor{gray}{(ICCV'23)}} & \ding{55} & \textbf{48.3} & \textbf{52.8} & $43.8$ & $48.8$ & $45.1$ & $49.8$ & $45.7$ & $50.5$ & $48.1$  \\
    MAI{$^\dagger$}~\cite{mai} \footnotesize{\textcolor{gray}{(ICLR'25)}} & \ding{55} & $35.0$ & $43.0$ & $31.8$ & $39.0$ & $43.5$ & $51.0$ & $36.8$ & $44.3$ & $40.6$  \\
    \hline
    ImageScope{$^\dagger$}~\cite{imagescope} \footnotesize{\textcolor{gray}{(WWW'25)}} & \ding{51} & $22.9$ & $28.4$ & $41.6$ & $47.0$ & $44.8$ & $51.6$ & $36.4$ & $42.3$ & $39.4$  \\
    \hline
    \rowcolor{blue!8}
    UniCVR (Stage~I) & \ding{51} & $39.5$ & $47.0$ & \underline{$54.6$} & \underline{$61.5$} & \underline{$58.8$} & \underline{$65.1$} & \underline{$51.0$} & \underline{$57.9$} & \underline{$54.4$}  \\
    \rowcolor{blue!8}
    UniCVR (Stage~II) & \ding{51} & \underline{$43.1$} & \underline{$49.5$} & \textbf{59.5} & \textbf{65.1} & \textbf{61.3} & \textbf{67.0} & \textbf{54.6} & \textbf{60.5} & \textbf{57.6}  \\
    \hline
    \end{tabular}
    }
\end{table*}

\begin{table}[t]
    \centering
    \caption{Performance comparison on WebVid-CoVR.}
    \label{tab:exp_web}
    \fit{
    \begin{tabular}{l|c|cccc|c}
    \hline
    Model & ZS & R@$1$ & R@$5$ & R@$10$ & R@$50$ & Avg. \\
    \hline \hline
    CoVR~\cite{covr} \footnotesize{\textcolor{gray}{(AAAI'24)}} & \ding{55} &
    $53.13$ & $79.93$ & $86.85$ & $97.69$ & $79.40$ \\
    CoVR\_Enrich~\cite{covrenrich} \footnotesize{\textcolor{gray}{(CVPR'24)}} & \ding{55} &
    $60.12$ & $84.32$ & $91.27$ & \underline{$98.72$} & $83.61$ \\
    CoVR-2~\cite{covr2} \footnotesize{\textcolor{gray}{(TPAMI'24)}} & \ding{55} &
    $59.82$ & $83.84$ & $91.28$ & $98.24$ & $83.30$ \\
    FDCA~\cite{fdca} \footnotesize{\textcolor{gray}{(ICLR'25)}} & \ding{55} &
    $54.80$ & $82.27$ & $89.84$ & $97.70$ & $81.15$ \\
    HUD~\cite{hud} \footnotesize{\textcolor{gray}{(MM'25)}} & \ding{55} &
    \underline{$63.38$} & \textbf{86.93} & \textbf{92.29} & \textbf{98.76} & \underline{$85.34$} \\
    \hline
    CoVR~\cite{covr} \footnotesize{\textcolor{gray}{(AAAI'24)}} & \ding{51} &
    $45.46$ & $70.46$ & $79.54$ & $93.27$ & $ 72.18 $ \\
    CoVR-2~\cite{covr2} \footnotesize{\textcolor{gray}{(TPAMI'24)}} & \ding{51} &
    $45.66$ & $71.71$ & $81.30$ & $94.80$ & $ 73.37 $ \\
    MoRe~\cite{more} \footnotesize{\textcolor{gray}{(CVPR'26)}} & \ding{51} &
    $63.00$ & $83.40$ & $87.60$ & - & - \\
    \hline
    \rowcolor{blue!8}
    UniCVR (Stage~I) & \ding{51} &
    $62.96$ & $83.67$ & $90.37$ & $97.81$ & $ 83.70 $ \\
    \rowcolor{blue!8}
    UniCVR (Stage~II) & \ding{51} &
    \textbf{66.77} & \underline{$86.31$} & \underline{$91.88$} & $98.29$ & \textbf{85.81} \\
    \hline
    \end{tabular}
    }
\end{table}

\textbf{(1) UniCVR achieves cutting-edge performance across all three CVR tasks.} Even without MLLM-guided dual-level reranking, Stage~I alone already outperforms most baselines. After Stage~II, UniCVR further advances state-of-the-art across all benchmarks (especially on CIRCO with $43.09\%$ mAP@5 vs.\ $34.10\%$ of MagicLens), and even rivals supervised methods on MT-FashionIQ (\eg, $57.6\%$ Avg.\ vs.\ $48.1\%$ of the FashionNTM) and WebVid-CoVR (\eg, $85.81\%$ Avg.\ vs.\ $85.34\%$ of the HUD). These results validate the effectiveness of both stages: Stage~I establishes a strong zero-shot foundation via MLLM-VLP alignment pre-training, while Stage~II further refines retrieval through MLLM-guided dual-level reranking, together enabling a unified framework that generalizes well across diverse CVR tasks.

\textbf{(2) Stage~II consistently improves over Stage~I, especially at top ranks.} The gains are observed across all five benchmarks: R@10 on FashionIQ (Avg.) improves by +3.63\% in absolute terms ($42.38\%{\to}46.01\%$), R@1 on CIRR by +7.83\% ($40.10\%{\to}47.93\%$), mAP@5 on CIRCO by +8.12\% ($34.97\%{\to}43.09\%$), R@5 on MT-FashionIQ (Avg.) by +3.6\% ($51.0\%{\to}54.6\%$), and R@1 on WebVid-CoVR by +3.81\% ($62.96\%{\to}66.77\%$). Meanwhile, the gains at R@50 are consistently modest (\eg, $+0.22\%$ on CIRR and $+0.48\%$ on WebVid-CoVR), which indicates that Stage~I already places most relevant targets within the top-50 and that Stage~II mainly reorders the head of the ranking, pushing targets from mid positions into the very top. The particularly large gain on CIRCO further suggests that the MLLM-guided dual-level reranking is especially effective at disambiguating among multiple plausible ground truths.

\textbf{(3) Treating the MLLM as a query embedder generalizes better than using it as a captioner.} Training-free methods such as LDRE, CIReVL, OSrCIR, AutoCIR, CoTMR, and G-MIXER leverage powerful MLLMs, either closed-source (\eg, GPT series) or large open-source models (\eg, Qwen2-VL-72B), as query reasoners to generate target descriptions. While these methods benefit from strong multimodal reasoning, they exhibit inconsistent cross-dataset performance. For instance, on FashionIQ (Avg.) G-MIXER and CoTMR trail our UniCVR (Stage~II) by $4.32$ and $5.38$ points ($50.85$ and $49.79$ vs. $55.17$), yet this gap widens to $11.30$ and $10.86$ points in mAP@5 on CIRCO ($31.79$ and $32.23$ vs. $43.09$).
By contrast, UniCVR achieves top performance consistently across all benchmarks without any task-specific prompts, demonstrating that bypassing the query-to-text bottleneck improves generalization across diverse downstream tasks. Notably, this is achieved with a query reasoner that is \emph{smaller} than those of several baselines, indicating that the advantage lies in how the MLLM is used rather than in its capacity.

\textbf{(4) Multi-source pre-training yields better generalization than any single data source.} Pre-training-based methods (\eg, FiRE, MagicLens, and CompoDiff) exemplify the limitation of relying on a single data source. FiRE, pre-trained solely on Type~II data (FiGMaQ) with pseudo modification texts, suffers from limited modification diversity and lacks domain-specific coverage (\eg, fashion), resulting in strong performance on general-domain benchmarks like CIRR and CIRCO but noticeably weaker rankings on FashionIQ.
More strikingly, even scaling up a single data source does not resolve this issue: MagicLens and CompoDiff are trained on $\sim$36.7M and $\sim$18.8M pseudo triplets, respectively, both far exceeding our $\sim$3.5M samples, yet neither achieves consistent top-tier rankings across benchmarks. By contrast, UniCVR Stage~I combines complementary Type~I, II, and III data, achieving consistently strong rankings across all five benchmarks. This confirms that given the inevitable limitations of any single data source, diversity matters more than scale alone in the zero-shot pre-training stage.

\subsection{Ablation Study}
\label{ssec:ablation}
To verify the effectiveness of each component in UniCVR, we compare the full model with its following variants.

\begin{table*}[ht]
    \centering
    \caption{Ablation study results. For FashionIQ and MT-FashionIQ, we report the average across three categories. For CIRR, we report the average of R@1/5/10/50 and R\textsubscript{subset}@1/2/3. For WebVid-CoVR, we report the average of R@1/5/10/50. Avg.\ is the mean of all individual metrics.}
    \label{tab:ablation}
    \fit{
    \begin{tabular}{l|cc|cc|cc|cc|c|c}
    \hline
    \multirow{2}{*}{Model} &  \multicolumn{2}{c|}{FashionIQ} & \multicolumn{2}{c|}{CIRR} & \multicolumn{2}{c|}{CIRCO} & \multicolumn{2}{c|}{MT-FashionIQ} & WebVid-CoVR & \multirow{2}{*}{Avg.} \\
    \cline{2-10}
    & R@$10$ & R@$50$ & R@$k$ & R\textsubscript{subset}@$k$ & mAP@$5$ & mAP@$10$ & R@$5$ & R@$8$ & Avg. R@$k$ & \\
    \hline \hline
    Stage~I (4B--PEcore) & $38.67$ & $60.70$ & $68.13$ & $80.46$ & $25.66$ & $27.42$ & $46.35$ & $53.77$ & $81.65$ & $53.65$ \\
    Stage~I (4B--CLIP-G) & $36.53$ & $58.13$ & $66.79$ & $78.80$ & $22.43$ & $24.04$ & $45.66$ & $51.66$ & $72.15$ & $50.69$  \\
    Stage~I (4B--4B) & $31.38$ & $52.04$ & $67.04$ & $80.09$ & $21.56$ & $22.87$ & $36.97$ & $44.11$ & $80.67$ & $48.53$  \\
    Stage~I (8B--PEcore) & $41.14$ & $62.21$ & $70.68$ & $81.90$ & $24.61$ & $26.51$ & $49.86$ & $56.92$ & $84.05$ & $55.32$ \\
    Stage~I w/o-Clustering & $42.33$ & $64.51$ & $70.01$ & $80.31$ & $31.56$ & $32.94$ & $50.81$ & $58.39$ & $84.37$ & $57.25$ \\
    \rowcolor{blue!8}
    Stage~I (Ours) & $42.38$ & $63.85$ & $72.10$ & $82.71$ & $34.97$ & $36.40$ & $50.99$ & $57.87$ & $83.70$ & $58.33$ \\
    \hline
    Stage~II w/-LocalOnly & $45.36$ & $64.00$ & $75.33$ & $84.98$ & $39.89$ & $41.30$ & $53.58$ & $60.24$ & $85.06$ & $61.08$ \\
    Stage~II w/-GlobalOnly & $44.36$ & $64.42$ & $ 74.54 $ & $83.97$ & $39.04$ & $40.24$ & $52.96$ & $59.48$ & $85.35$ & $60.48$ \\
    Stage~II w/-GlobalOnly\_{\text{sim}} & $42.44$ & $63.52$ & $71.61$ & $81.98$ & $34.78$ & $36.32$ & $50.86$ & $57.72$ & $83.24$ & $58.05$ \\
    \rowcolor{blue!8}
    Stage~II (Ours)  & $46.01$ & $64.32$ & $76.10$ & $85.41$ & $43.09$ & $43.86$ & $54.64$ & $60.53$ & $85.81$ & $62.20$ \\
    \hline
    \end{tabular}
    }
\end{table*}

\begin{itemize}[leftmargin=*]
\item \textbf{Query-encoder scale (Stage~I (4B--PEcore) / Stage~I (8B--PEcore)).} To investigate the effect of the query encoder's capacity, we fix the target to the frozen PEcore-G and replace the default Qwen3-VL-32B query encoder with Qwen3-VL-4B and Qwen3-VL-8B, obtaining \textbf{Stage~I (4B--PEcore)} and \textbf{Stage~I (8B--PEcore)}, respectively.

\item \textbf{Target encoder (Stage~I (4B--CLIP-G) / Stage~I (4B--4B)).} To examine our asymmetric MLLM--VLP design against the alternatives, we fix the query encoder to 4B (which keeps the cost low) and vary only the target side. \textbf{Stage~I (4B--4B)} replaces the frozen VLP target with a trainable Qwen3-VL-4B that shares parameters with the query encoder, forming the symmetric MLLM--MLLM design; \textbf{Stage~I (4B--CLIP-G)} keeps a frozen VLP target but swaps PEcore-G (1.9B) for a comparably sized CLIP-G (1.84B). The former tests whether the target should be a frozen VLP at all, and the latter why we choose PEcore-G over the widely used CLIP.

\item \textbf{Stage~I w/o-Clustering:} To check the effect of the cluster-based hard negative sampling strategy, we remove the clustering step while still ensuring each mini-batch is drawn from a single data source, still enabling effective contrastive learning across multiple sources.

\item \textbf{Stage~II w/-LocalOnly:} To isolate the contribution of local re-scoring, we skip the global query refinement step and directly fuse the MLLM relevance scores with the initial Stage~I cosine similarities for candidates within the scored subset, while leaving the remaining candidates unchanged.

\item \textbf{Stage~II w/-GlobalOnly:} Similarly, to isolate the contribution of global re-scoring, we skip the local re-scoring in MLLM-guided dual-level reranking.

\item \textbf{Stage~II w/-GlobalOnly}\_{\textbf{sim}}: To verify the importance of using MLLM relevance scores for query refinement, we replace the MLLM-derived scores $s_g$ in Eq.~\eqref{eq:expansion} with the Stage~I cosine similarities for weighting the candidate embeddings, while keeping the rest of the global re-scoring pipeline unchanged.
\end{itemize}

Table~\ref{tab:ablation} reports the ablation results on all five benchmarks, from which we draw the following findings.

\textbf{(1) Query-encoder scale.}
Both Stage~I (4B--PEcore) and Stage~I (8B--PEcore) underperform Stage~I (Ours), demonstrating that scaling up the MLLM query encoder generally strengthens compositional query understanding. However, the benefit varies across benchmarks. The improvement is particularly pronounced on the general-domain CIRCO and CIRR benchmarks: increasing the encoder size from 4B to 32B improves mAP@5 on CIRCO from $25.66\%$ to $34.97\%$, and R@$k$ on CIRR by an average of $3.97$ percentage points. These benchmarks contain more diverse visual concepts and modification intents, while CIRCO additionally provides multiple semantically valid targets, demanding finer compositional reasoning and visual disambiguation. In contrast, the improvements on FashionIQ and MT-FashionIQ are more moderate. As both benchmarks are restricted to the fashion domain with relatively regular object categories, visual structures, and modification patterns, the smaller encoders already capture much of the required compositional information. On WebVid-CoVR, the 8B model even slightly outperforms the 32B model. Overall, larger query encoders are most beneficial for general-domain benchmarks involving diverse visual interpretations and fine-grained compositional distinctions, whereas smaller encoders remain competitive on domain-constrained benchmarks with more regular data distributions.

\textbf{(2) Target encoder.}
The target encoder variants isolate our design choices on the target side. First, replacing the frozen VLP target with a parameter-shared trainable MLLM (Stage~I (4B--4B)) degrades performance against Stage~I (4B--PEcore). While prevailing universal multimodal retrievers~\cite{mmembed,gme} adopt exactly this symmetric MLLM--MLLM design, they rely on large-scale, high-quality supervision, which is unavailable in our zero-shot setting where the pre-training triplets are auto-curated and inevitably noisy; a trainable target then lets this noise reshape the target space and erode the metric structure that retrieval relies on. Keeping the target as a frozen VLP with an already well-structured metric space instead preserves reliable alignment under such imperfect supervision. Second, swapping PEcore-G for a comparably sized CLIP-bigG (Stage~I (4B--CLIP-G)) keeps the image benchmarks close but collapses on WebVid-CoVR ($81.65\%$ vs.\ $72.15\%$), since CLIP lacks native video encoding; this motivates our choice of PEcore-G, whose unified image-video space enables a single gallery encoder across all three tasks.

\textbf{(3) Hard negative sampling.}
Stage~I (Ours) outperforms Stage~I w/o-Clustering, especially on CIRR ($72.10\%$ vs.\ $70.01\%$ R@$k$) and CIRCO ($34.97\%$ vs.\ $31.56\%$ mAP@5). This indicates that the cluster-based hard negative sampling strategy is effective for learning discriminative embeddings, particularly on general-domain benchmarks where semantically similar candidates are more prevalent.

\textbf{(4) Dual-level reranking.}
Both Stage~II w/-LocalOnly and Stage~II w/-GlobalOnly improve over Stage~I, but their combination in Stage~II (Ours) achieves the best performance. This validates the complementary nature of the two re-scoring levels: global re-scoring refines the query embedding to improve similarity estimation across the entire gallery, while local re-scoring directly leverages fine-grained MLLM relevance judgments within the scored subset. Stage~II w/-GlobalOnly\_sim, which replaces the MLLM relevance scores with Stage~I cosine similarities for query refinement, yields negligible improvement over Stage~I and performs substantially worse than Stage~II w/-GlobalOnly. This confirms that the MLLM provides substantially more reliable relevance signals than embedding similarities for identifying and weighting pseudo-positive candidates during query refinement.

\begin{table*}[ht]
    \centering
    \caption{Analysis on the pre-training data. The reported metrics follow Table~\ref{tab:ablation}. Within each metric column, the top three of the four data configurations are shaded from darkest (best) to lightest, while the ReAlign ablation is excluded from the ranking.}
    \label{tab:training-data}
    \fit{
    \begin{tabular}{l|cc|cc|cc|cc|c|c}
    \hline
    \multirow{2}{*}{Training Data} &  \multicolumn{2}{c|}{FashionIQ} & \multicolumn{2}{c|}{CIRR} & \multicolumn{2}{c|}{CIRCO} & \multicolumn{2}{c|}{MT-FashionIQ} & WebVid-CoVR & \multirow{2}{*}{Avg.} \\
    \cline{2-10}
    & R@$10$ & R@$50$ & R@$k$ & R\textsubscript{subset}@$k$ & mAP@$5$ & mAP@$10$ & R@$5$ & R@$8$ & Avg. R@$k$ & \\
    \hline \hline
    Type I Only & \cA{$29.01$} & \cA{$51.79$} & \cC{$55.20$} & \cC{$75.96$} & \cC{$12.68$} & \cC{$13.46$} & \cA{$37.00$} & \cA{$44.53$} & \cC{$75.57$} & \cB{$43.91$} \\
    \cellcolor{ablrow}Type I Only \textit{w/o} ReAlign & \cellcolor{ablrow}$28.59$ & \cellcolor{ablrow}$51.26$ & \cellcolor{ablrow}$54.75$ & \cellcolor{ablrow}$75.55$ & \cellcolor{ablrow}$12.20$ & \cellcolor{ablrow}$13.06$ & \cellcolor{ablrow}$36.61$ & \cellcolor{ablrow}$43.97$ & \cellcolor{ablrow}$75.37$ & \cellcolor{ablrow}$43.48$ \\
    Type II Only & \cC{$14.65$} & \cC{$32.56$} & \cB{$63.06$} & \cA{$80.07$} & \cA{$17.39$} & \cB{$18.36$} & \cC{$17.94$} & \cC{$22.09$} & $72.33$ & \cC{$37.61$} \\
    Type III Only & $10.41$ & $25.86$ & $50.20$ & $65.69$ & $4.49$ & $5.32$ & $16.45$ & $20.82$ & \cB{$75.82$} & $30.56$ \\
    Multi-Source (Ours) & \cB{$26.15$} & \cB{$49.32$} & \cA{$64.33$} & \cB{$79.71$} & \cB{$17.33$} & \cA{$18.43$} & \cB{$33.15$} & \cB{$41.06$} & \cA{$76.77$} & \cA{$45.14$} \\
    \hline
    \end{tabular}
    }
\end{table*}

\begin{figure*}[t]
    \centering

    \subfloat[LLaVA-Pretrain (general domain)]{%
        \includegraphics[width=0.32\textwidth]{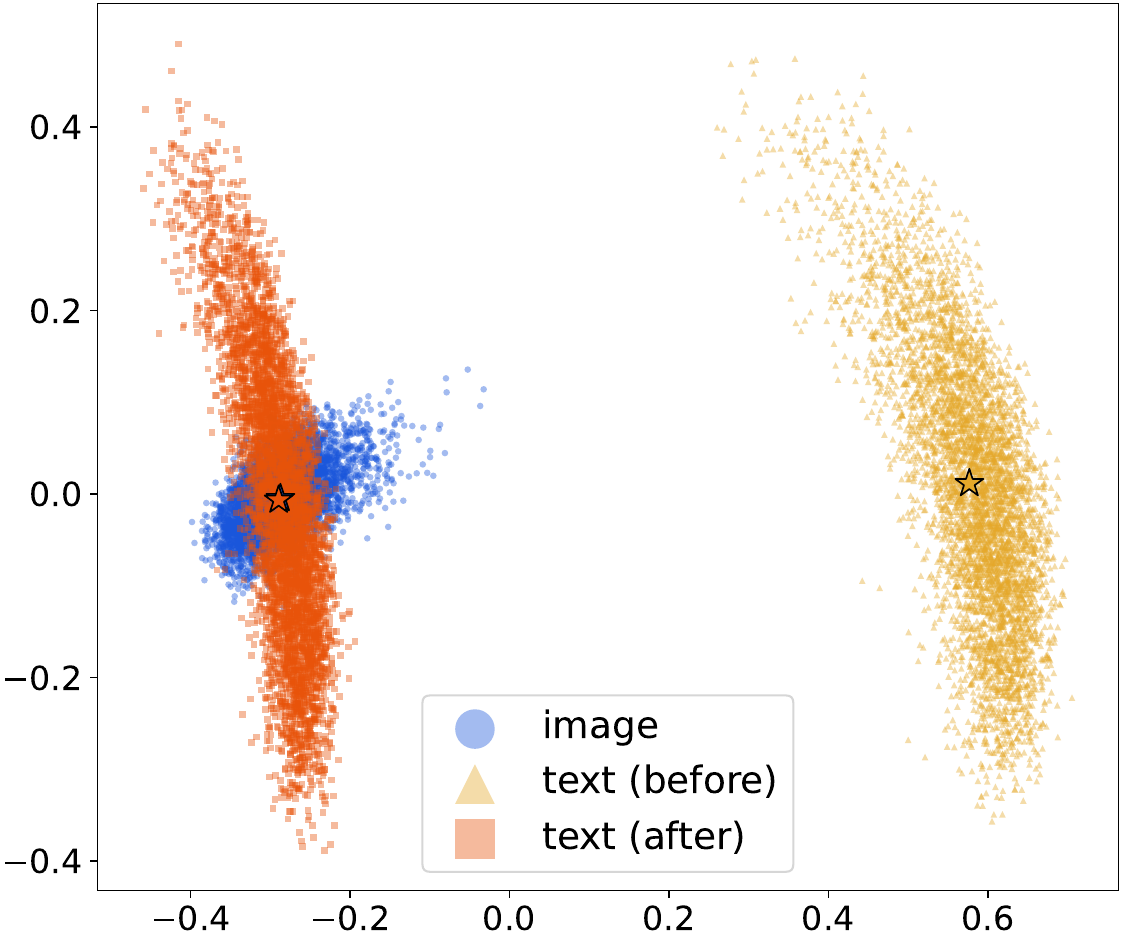}%
        \label{fig:realign_llava}%
    }
    \hspace{0.12\textwidth}
    \subfloat[Fashion200K (fashion domain)]{%
        \includegraphics[width=0.32\textwidth]{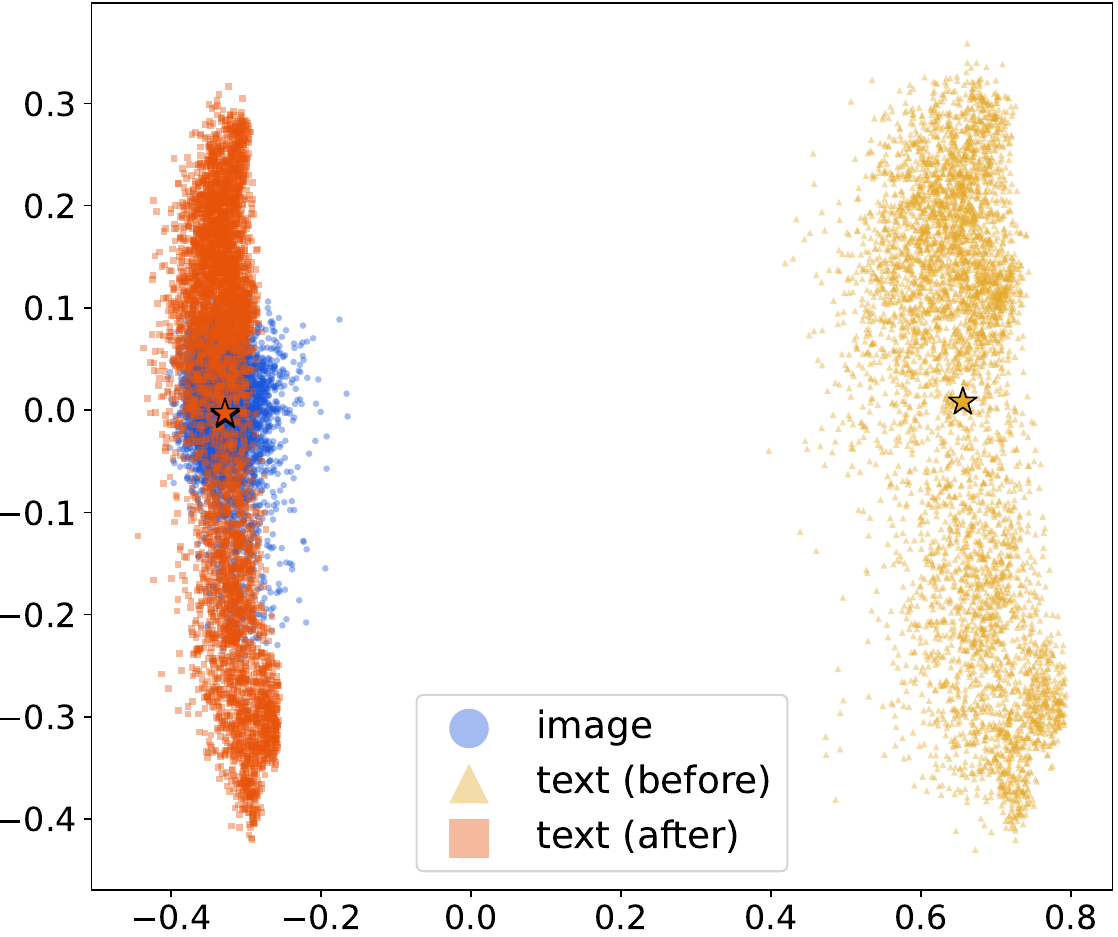}%
        \label{fig:realign_fashion}%
    }

    \caption{PCA visualization of the embedding distributions before and after the ReAlign transformation on (a) LLaVA-Pretrain and (b) Fashion200K. Yellow triangles and orange squares denote text embeddings before and after ReAlign, respectively; blue circles denote target image embeddings; and stars mark cluster centroids.}
    \label{fig:realign_pca}
\end{figure*}

\subsection{Analysis on the Pre-Training Data}
\label{sec:data-analysis}
Each of our three data types contributes a distinct kind of supervision, and none of them covers what the others provide. We test whether this complementarity actually translates into better zero-shot generalization by varying only the pre-training corpus.
For a controlled and computationally affordable comparison, all variants use the same architecture (Qwen3-VL-4B as the query encoder with the frozen PEcore-G gallery encoder), the same training budget (60K triplets), and the same optimization schedule (1,000 training steps). Consequently, any performance difference can be attributed solely to the training data. Specifically, we include the following variants: 
\begin{itemize}[leftmargin=*]
    \item \textbf{Type~I Only}: 30K triplets from LLaVA-Pretrain and 30K from Fashion200K, maintaining a balanced mix of general and fashion-domain data.
    \item \textbf{Type~I Only \textit{w/o} ReAlign}: Unlike Type~II and Type~III, Type~I uses text descriptions as retrieval targets, introducing a cross-modal gap. To examine whether ReAlign is necessary for mitigating this gap and effectively leveraging Type~I data, we include this variant, which is identical to \textbf{Type~I Only} except that it directly uses the PEcore text embedding of the target description as the retrieval target, without applying ReAlign.
    \item \textbf{Type~II Only} and \textbf{Type~III Only}: 60K triplets sampled from FiGMaQ and filtered AnyEdit, respectively.
    \item \textbf{Multi-Source}: The same 60K budget is distributed across all three data types, with 10K LLaVA-Pretrain, 10K Fashion200K, 20K FiGMaQ, and 20K AnyEdit, following the composition strategy of our full training corpus.
\end{itemize}
 
From Table~\ref{tab:training-data}, we obtained the following observations. \textbf{(1) Each single source transfers well only where its own characteristics happen to match.} Type~I Only is the only single source that stays competitive on both the fashion-domain and the general-domain benchmarks, attaining the best fashion-domain results ($29.01$ R@10 on FashionIQ and $37.00$ R@5 on MT-FashionIQ) and the second-best on CIRR and CIRCO. This is mainly because it is the only source that draws on a fashion corpus and a general-domain corpus at once, and its MLLM-generated modification texts span a far wider range of intents than those of the other two types. Type~II Only takes the top general-domain results instead ($63.06$ R@$k$ on CIRR and $17.39$ mAP@5 on CIRCO), since its real target images provide genuine visual matching supervision that the pseudo target descriptions of Type~I cannot, yet it collapses on fashion ($14.65$ R@10), as its modification texts are derived from naturally occurring differences between mined ImageNet pairs and rarely describe attribute-level garment edits. Type~III Only yields the lowest overall average ($30.56$). This is likely because its synthesized targets are directly edited from the source images and thus remain highly similar in visual appearance. Such data provides highly accurate instruction--target correspondence, but lacks diverse semantic and visual variations compared with Type~I and Type~II, which restricts the model's ability to handle broader retrieval scenarios. This characteristic also explains its superior performance on WebVid-CoVR, where the target clip typically differs from the reference through localized visual changes, closely matching the supervision characteristics of image-editing data.

\textbf{(2) Combining the three types outperforms any single source.}
Multi-Source achieves the best overall average and ranks first on most individual metrics, despite using only a fraction of the samples from each single-source setting. This indicates that the three data types provide complementary supervision rather than interchangeable signals, highlighting the importance of data diversity for zero-shot pre-training. Its only weakness appears on the two fashion benchmarks, where it underperforms Type~I Only ($26.15$ vs. $29.01$ R@10), mainly because it uses fewer fashion-domain samples (10K vs. 30K).

\textbf{(3) ReAlign provides consistent gains.}
Removing ReAlign from Type~I Only degrades performance on all metrics, reducing the average from $43.91$ to $43.48$. The consistent improvement across three tasks demonstrates the effectiveness of ReAlign in bridging the modality gap between the text-form targets in Type~I data and the image gallery used at test time. Since ReAlign is computed offline from fixed statistics and introduces no additional training cost, we adopt it for all Type~I samples by default.

We further visualize the effect of ReAlign on the embedding distributions, so as to verify that the transformation does close the modality gap it targets. For each Type~I source we randomly sample 5,000 descriptions and 5,000 images, encode them with the frozen PEcore text and image encoders, additionally apply ReAlign to the text embeddings, and project all of them into a shared 2D space via PCA. As shown in Figure~\ref{fig:realign_pca}, before ReAlign the text and image embeddings form two clearly separated clusters with distant centroids, which is the discrepancy a Type~I-trained query encoder would inherit, since it is optimized against text-form targets but evaluated against an image gallery. After ReAlign, the transformed text embeddings shift onto the image manifold and overlap substantially with the image distribution, with nearly coincident centroids. Note that the sampled texts and images are not paired, so exact coincidence is neither expected nor required. What matters is that the systematic offset between the two modalities is removed.

\subsection{Efficiency Analysis}
\label{ssec:efficiency}
Retrieval efficiency is critical in real-world applications. We analyze the efficiency of our UniCVR here: we first establish that a small scoring budget suffices, so that its cost stays bounded regardless of gallery size, and then quantify the per-query cost of both stages.

\begin{figure*}[t]
    \centering

    \begin{minipage}[c]{0.36\textwidth}
        \centering
        \includegraphics[width=\linewidth]{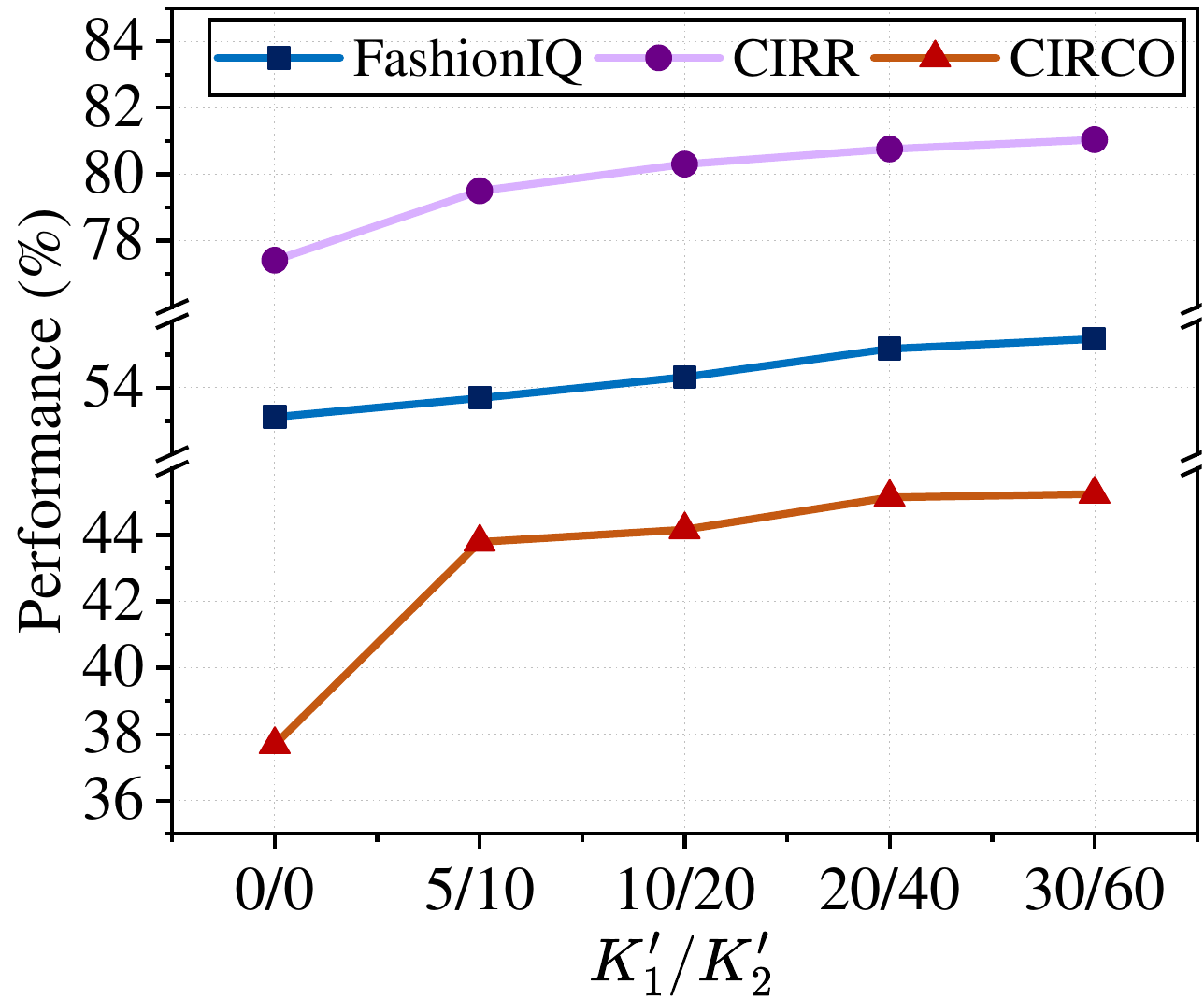}
    \end{minipage}
    \hfill
    \begin{minipage}[c]{0.56\textwidth}
        \centering

        \setlength{\tabcolsep}{7pt}
        \renewcommand{\arraystretch}{1.35}

        \begin{tabular}{lccc}
            \toprule
            $K'_1/K'_2$ & FashionIQ & CIRR & CIRCO \\[0.3em]
            \midrule
            $0/0$   & --       & --       & --       \\
            $5/10$  & $52.2\%$ & $37.1\%$ & $33.9\%$ \\
            $10/20$ & $57.2\%$ & $40.4\%$ & $37.9\%$ \\
            $20/40$ & $60.9\%$ & $41.7\%$ & $40.8\%$ \\
            $30/60$ & $62.4\%$ & $42.5\%$ & $42.0\%$ \\
            \bottomrule
        \end{tabular}

        \vspace{1em}

        \begin{minipage}{0.96\linewidth}
            \raggedright
            \footnotesize
            \textit{Note:} The values denote the percentages of queries
            terminated at $K'_1$ without extending to $K'_2$.
        \end{minipage}
    \end{minipage}

    \caption{Effect of the scoring budget $K'_1/K'_2$.
    Performance denotes the average of all reported metrics on each benchmark.
    The table reports the early termination ratio under different configurations.}
    \label{fig:k1k2}
\end{figure*}

\textbf{A small budget suffices.}
Figure~\ref{fig:k1k2} reports the effect of the scoring budget $K'_1/K'_2$, together with the proportion of queries that terminate early at $K'_1$. Performance improves steadily from $5/10$ to $20/40$ and saturates beyond that, indicating that a moderate budget suffices for effective reranking; this is the empirical basis for our default of $20/40$, which sits on a plateau rather than at a tuned peak. Even the smallest budget $5/10$ already yields substantial gains, reflecting the fact that Stage~I places most relevant targets close to the head of the ranking. Crucially, because performance saturates at a small $K'$, the budget need not grow with the gallery, which is what keeps the cost of Stage~II bounded. The early-termination column shows that a substantial proportion of queries stop at $K'_1$: FashionIQ exhibits the highest ratio, as its targets are visually distinctive and easy for the assessor to verify among top-ranked candidates, whereas CIRCO, whose queries admit multiple plausible ground truths, more often requires the extended budget.

\textbf{Per-query cost.}
Table~\ref{tab:eff_compare} compares the per-query inference cost of UniCVR against the strongest training-free baselines. Gallery embeddings are pre-computed offline, and Stage~II reuses the same Qwen3-VL-32B with LoRA disabled. Stage~I is remarkably efficient, running at $0.16$\,s per query with no autoregressive generation and no external API call, yet it already matches or exceeds the strongest baselines on most benchmarks, delivering both high speed and strong accuracy.
By contrast, CoTMR and G-MIXER, the two strongest competitors on ZS-CIR tasks, both incur inherently heavier inference despite requiring no training, since each performs autoregressive (M)LLM generation before retrieval even begins. CoTMR runs two separate Qwen2-VL-72B chain-of-thought passes per query, with each reasoning pass alone taking over $3.18$\,s, an order of magnitude above Stage~I's single forward pass. G-MIXER issues a GPT-4o generation to produce its target description and then performs 7 mixup-ratio retrieval passes on top of it, and it further depends on a closed-source API that incurs per-call cost and offers no reproducibility guarantee.
Stage~II performs fine-grained cross-modal verification and is inevitably heavier, but this cost buys a substantial accuracy gain over Stage~I across all three benchmarks (\eg, CIRCO mAP@5 from $34.97$ to $43.09$). Moreover, the early-termination mechanism reduces its cost from $7.1$\,s to $4.9$\,s, a $31\%$ latency saving with no loss in accuracy, keeping the reranking overhead practical.

\begin{table*}[ht]
    \centering
    \caption{Efficiency comparison with the strongest training-free ZS-CIR baselines. Total Params sums the reasoner and retriever. Baseline timings are quoted from the respective papers under different settings, serving as order-of-magnitude references rather than a strictly controlled comparison.}
    \label{tab:eff_compare}
    \fit{
    \begin{tabular}{l|cc|c|cc}
    \hline
    Model & Reasoner & Retriever & Total Params & Autoreg.\ Gen. & Time \\
    \hline \hline
    CoTMR   & Qwen2-VL-72B    & CLIP ViT-G (1.84\,B) & $73.8$\,B & \checkmark & $>3.18$\,s \\
    G-MIXER & GPT-4o (closed) & CLIP ViT-G (1.84\,B) & --        & \checkmark & $\sim0.94$\,s \\
    UniCVR (Stage~I)  & Qwen3-VL-32B & PEcore-G (1.9\,B) & $33.9$\,B & $\times$ & $0.16$\,s \\
    \hline
    UniCVR (Stage~II, fixed $K'{=}40$) & Qwen3-VL-32B & PEcore-G (1.9\,B) & $33.9$\,B & $\times$ & $7.10$\,s \\
    UniCVR (Stage~II, adaptive) & Qwen3-VL-32B & PEcore-G (1.9\,B) & $33.9$\,B & $\times$ & $4.90$\,s \\
    \hline
    \end{tabular}
    }
\end{table*}

\subsection{Qualitative Analysis}
\label{ssec:qualitative}

\begin{figure}[!ht]
\centering
\includegraphics[width=\linewidth]{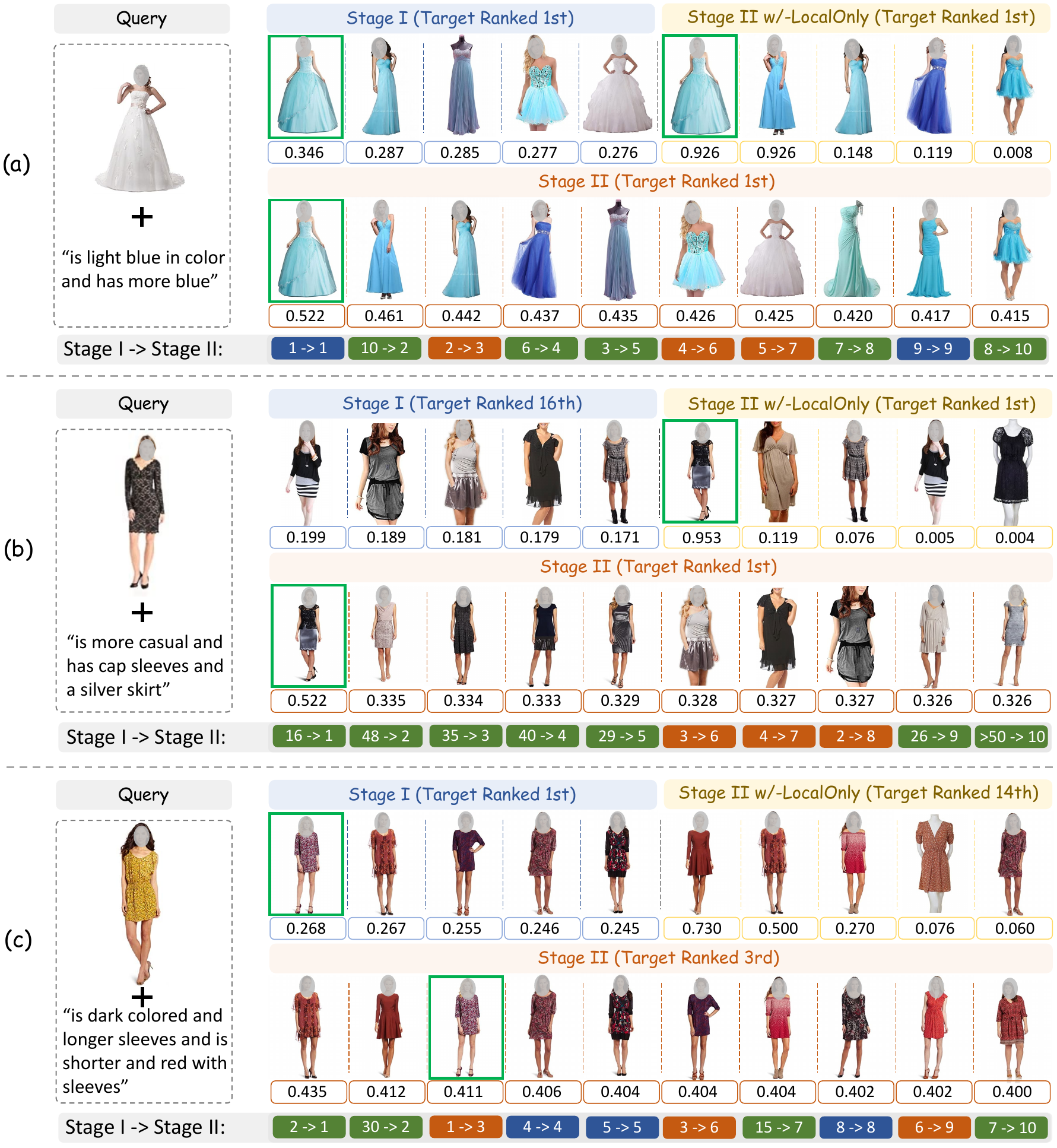}
\caption{Case studies on FashionIQ. Each case shows the composed query (left), the Stage~I ranking with cosine similarities (top-5 results), the Stage~II w/-LocalOnly ranking with MLLM relevance scores (top-5 results), and the final Stage~II ranking with dual-level re-scored values (top-10 results). Target images are highlighted in green boxes. Below the Stage~II results, rank changes from Stage~I$\to$Stage~II are annotated via background color: green = improvement, blue = unchanged, brown = demotion.}
\label{fig:case_fashioniq}
\end{figure}

\begin{figure}[!ht]
\centering
\includegraphics[width=\linewidth]{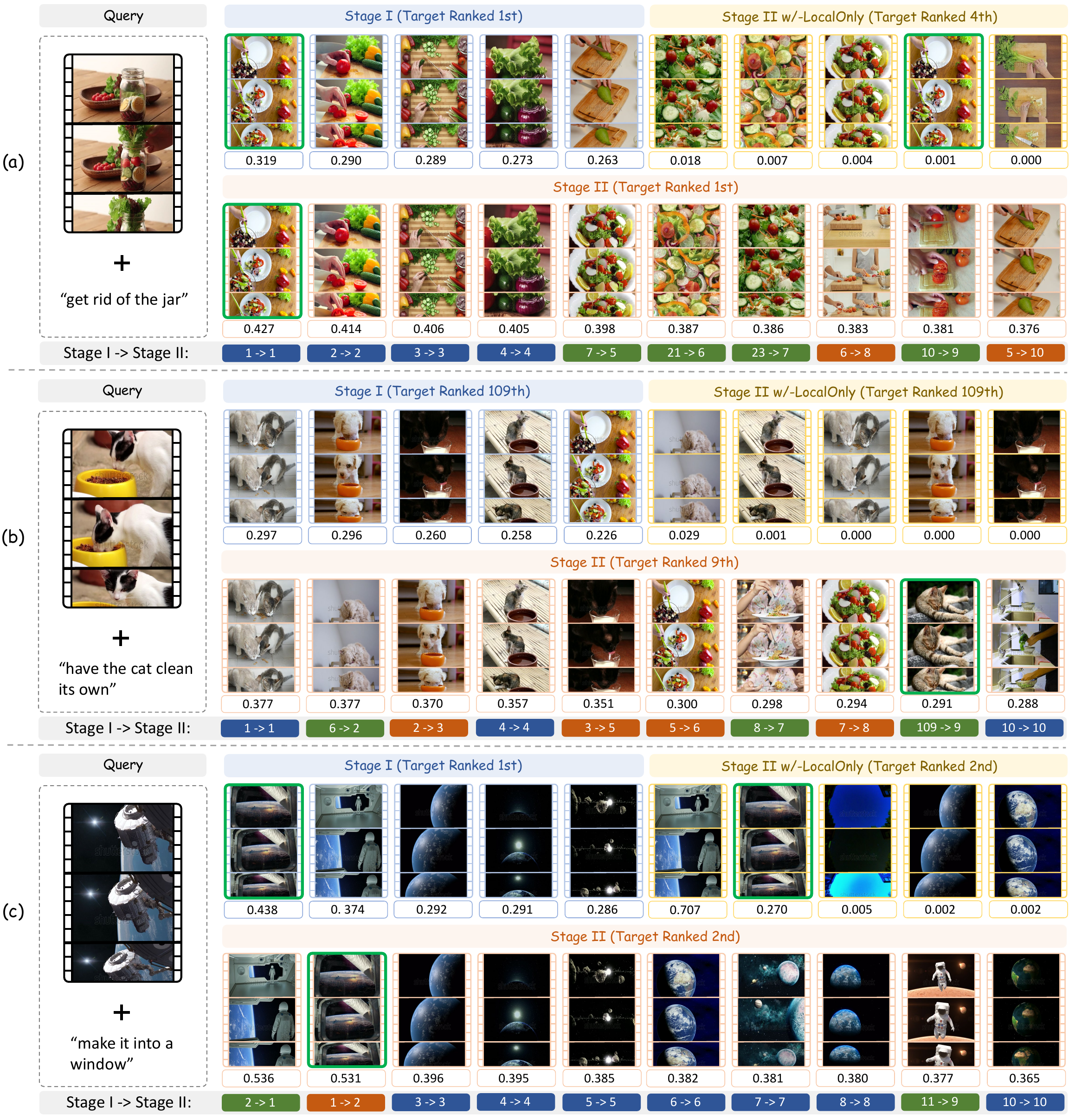}
\caption{Case studies on WebVid-CoVR. The layout follows the same format as Figure~\ref{fig:case_fashioniq}.}
\label{fig:case_webvid}
\end{figure}

To provide an intuitive understanding of how each stage contributes to the final ranking, we present case studies on FashionIQ (Figure~\ref{fig:case_fashioniq}) and WebVid-CoVR (Figure~\ref{fig:case_webvid}). These two benchmarks are chosen because CIRR and CIRCO require server-side evaluation and do not release ground-truth labels, because FashionIQ is the most widely used fashion benchmark and shares its data source with Multi-Turn FashionIQ, and because WebVid-CoVR provides a complementary open-domain video scenario. We highlight the following observations.

\textbf{(1) Stage~II preserves correct Stage~I rankings.} In Figure~\ref{fig:case_fashioniq}(a), the query asks for a dress that is light blue with more blue. Stage~I ranks the target at \#1 with a cosine similarity of $0.346$, and the assessor assigns a high score ($0.926$) to both the target and a visually similar candidate, confirming its relevance. After dual-level re-scoring the target stays at \#1, with only mild reordering among nearby candidates ($10\to2$, $2\to3$). A complementary case appears in Figure~\ref{fig:case_webvid}(a), where Stage~I again returns the correct target at \#1 for the query ``get rid of the jar'', but the assessor finds no confident match (the highest score is $0.018$). Stage~II then leaves the ordering essentially untouched. Reranking is thus conservative in both regimes, whether the initial ranking is confidently confirmed or not confidently contested, which matters because an aggressive reranker can easily destroy a head that was already correct.

\textbf{(2) Stage~II promotes initially under-ranked targets.} Figure~\ref{fig:case_fashioniq}(b) presents a harder query requesting a dress that is more casual with cap sleeves and a silver skirt. Stage~I places the target at \#16, deep enough that R@10 would miss it, but still inside the scoring budget. The assessor scores it $0.953$, far above the other assessed candidates, and dual-level re-scoring lifts it to \#1. This is the mechanism behind the concentration of Stage~II gains at top ranks observed in Section~\ref{ssec:comparison}.

\textbf{(3) Global re-scoring rescues targets beyond the scoring window.} In Figure~\ref{fig:case_webvid}(b), for the query ``have the cat clean its own'', Stage~I ranks the target at \#109, far outside the budget $K'_2=40$. The assessor therefore never sees it, and the LocalOnly variant leaves it at \#109. Global re-scoring, however, uses the scores of the \emph{assessed} candidates to refine the query embedding, and recomputing similarities over the full gallery with the refined embedding promotes the target from \#109 to \#9. This is the clearest evidence that the global level is not redundant with the local one, as only the global level can move items the MLLM never scored.

\textbf{(4) Demotions trace to incomplete annotations rather than errors.} In Figure~\ref{fig:case_fashioniq}(c), the target drops from \#1 to \#3. Inspecting the two promoted candidates shows that both are plausible matches for the modification but are simply not annotated as ground truths; the assessor is, in effect, penalized for being right. The same pattern recurs in Figure~\ref{fig:case_webvid}(c), where the target falls from \#1 to \#2, displaced by a candidate depicting a comparable window-like transformation that is likewise unlabeled. This is a known limitation of CIR benchmarks with incomplete annotations, and it implies that our reported numbers, particularly on single-ground-truth benchmarks such as FashionIQ and CIRR, understate the true effect of Stage~II.

\section{Conclusion}

In this paper, we presented UniCVR, a unified zero-shot framework for composed visual retrieval that jointly addresses CIR, MT-CIR, and CoVR without any task-specific human-annotated triplets. UniCVR combines the compositional reasoning of MLLMs with the well-structured metric space of VLP models through a two-stage design. In Stage~I, we bridge the heterogeneous embedding spaces between the MLLM query encoder and the frozen VLP gallery encoder via contrastive pre-training on a curated multi-source dataset, enhanced by a cluster-based hard negative sampling strategy. In Stage~II, an MLLM-guided dual-level reranking mechanism refines the initial retrieval results through adaptive budgeted subset scoring and dual-level re-scoring. Extensive experiments across five benchmarks demonstrate that UniCVR achieves cutting-edge performance on all three CVR tasks, confirming its effectiveness and generalizability. We hope UniCVR provides a foundation for future ZS-CVR research. In future work, we plan to explore more efficient MLLM reranking strategies and extend UniCVR to broader scenarios.

\bibliographystyle{IEEEtran}
\bibliography{sample-base}

\begin{IEEEbiography}[{\includegraphics[width=1in,height=1.25in,clip,keepaspectratio]{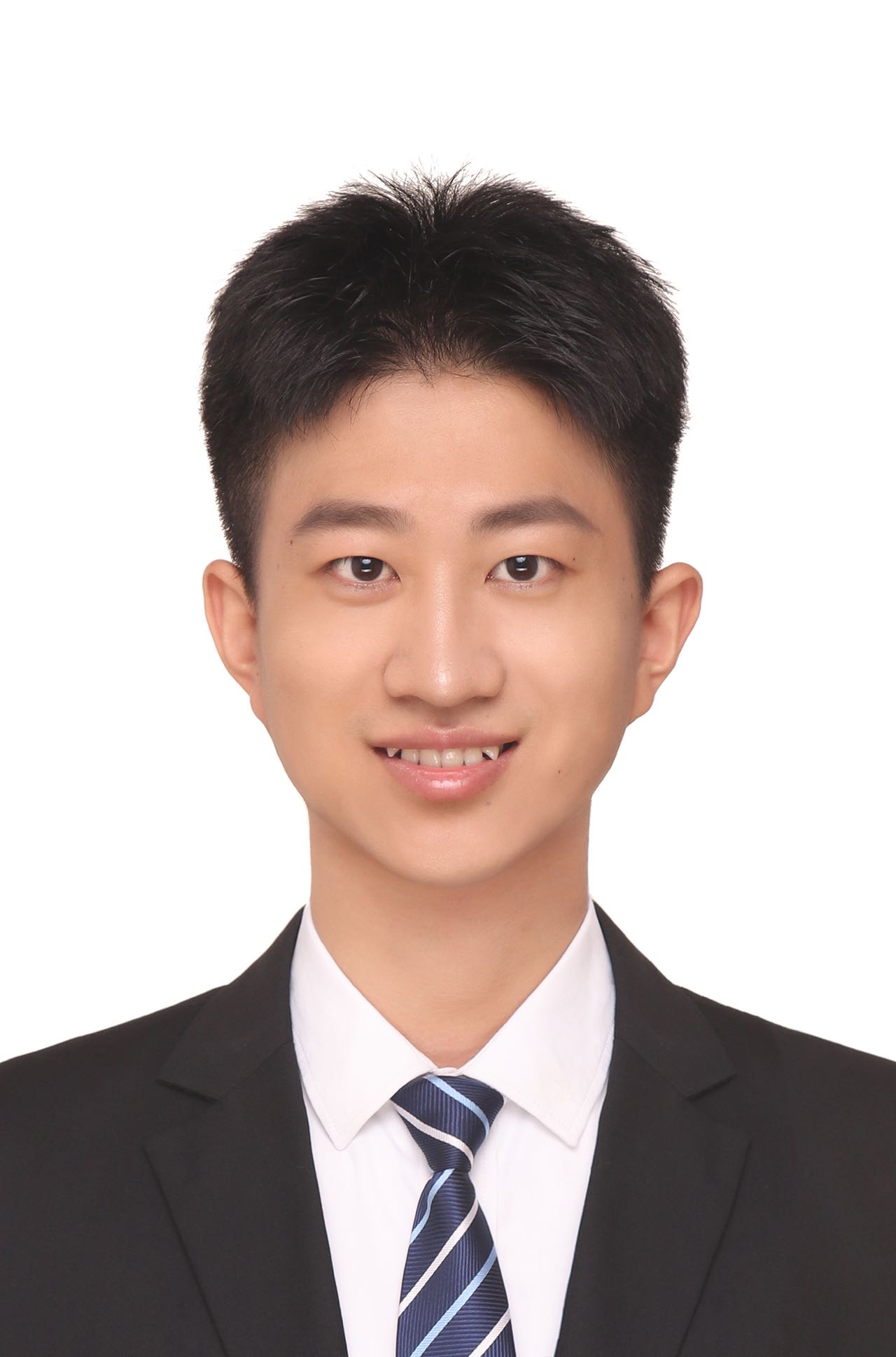}}]
{Haokun Wen} received the B.E. degree from Ocean University of China in 2019 and the M.S. degree from Shandong University in 2022. He is currently pursuing the Ph.D. degree with the School of Computer Science and Technology, Harbin Institute of Technology (Shenzhen), Shenzhen, China. His research interests include multimedia computing and information retrieval. He has published several papers in top venues such as ACM SIGIR, IEEE TIP, and IEEE TPAMI.
\end{IEEEbiography}

\begin{IEEEbiography}
[{\includegraphics[width=1in,height=1.25in,clip,keepaspectratio]{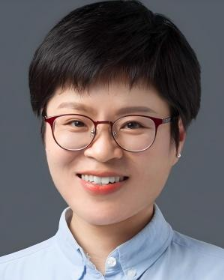}}]
{Xuemeng Song} (Senior Member, IEEE) received the B.E. degree from the University of Science and Technology of China, in 2012, and the Ph.D. degree from the School of Computing, National University of Singapore, in 2016. She is currently an Associate Professor with Southern University of Science and Technology. She has published several papers in the
top venues, such as ACM SIGIR, MM, and TOIS. She has served as a reviewer for many top conferences and journals. Her research interests include information retrieval and social network analysis. She is an Associate Editor of IET Image Processing and IEEE TRANSACTIONS ON CIRCUITS AND SYSTEMS FOR VIDEO TECHNOLOGY.
\end{IEEEbiography}

\begin{IEEEbiography}[{\includegraphics[width=1in,height=1.25in,clip,keepaspectratio]{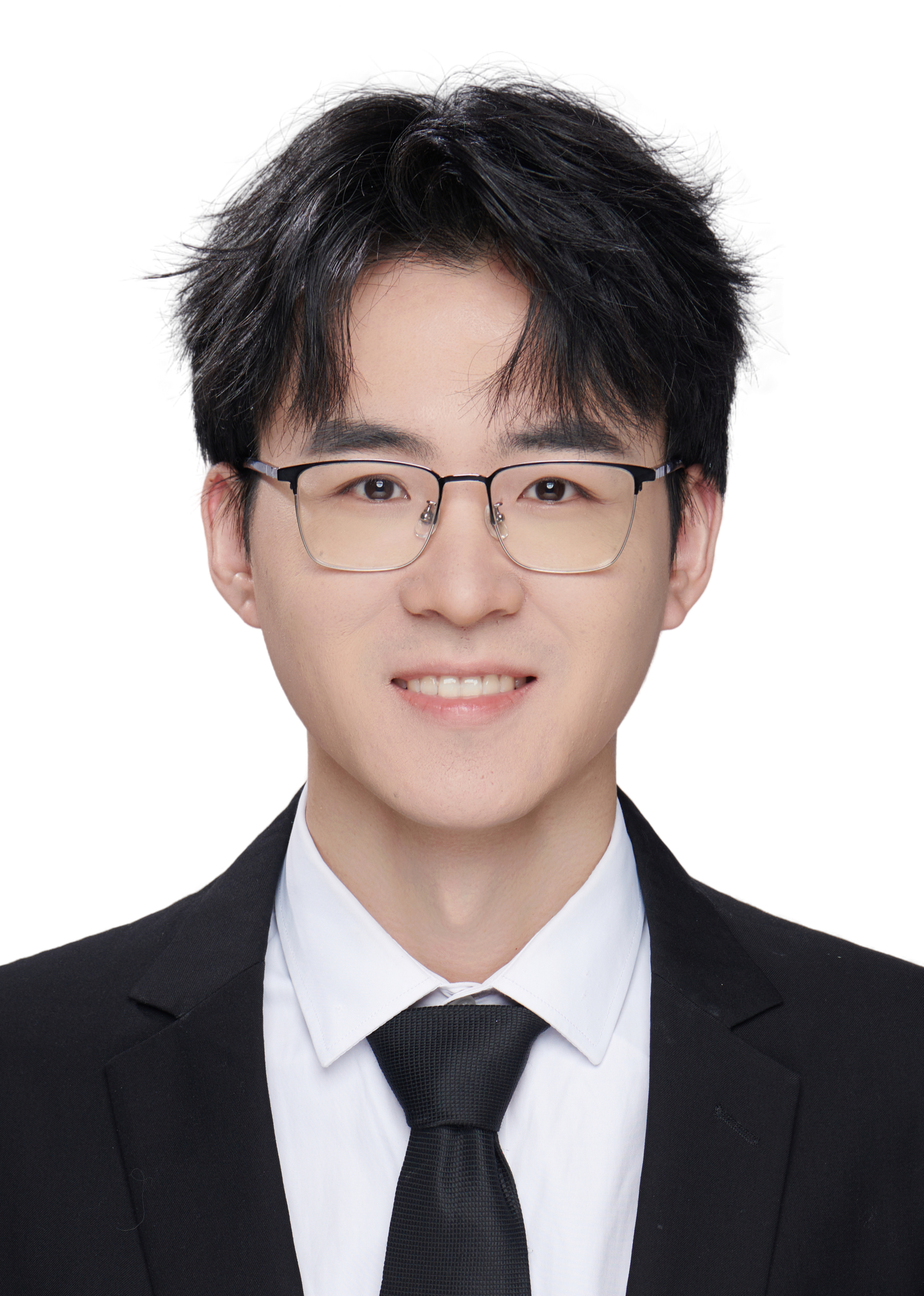}}]
{Haoyu Zhang} received the M.S. degree from Shandong University, in 2023. He is currently working toward the doctor's degree with the School of Computer Science and Technology, Harbin Institute of Technology (Shenzhen). His research interests include egocentric vision and spatial understanding. His work has been published in several top-tier conferences and journals, including IEEE TPAMI, CVPR, NeurIPS, ICML, AAAI, and ACM MM. He has served as a Reviewer for various conferences and journals, such as CVPR, ICCV, NeurIPS, ICML, ICLR, IEEE TPAMI, IEEE TKDE, and IEEE TMM.
\end{IEEEbiography}

\begin{IEEEbiography}
[{\includegraphics[width=1in,height=1.25in,clip,keepaspectratio]{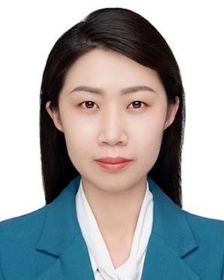}}]
{Weili Guan} (Member, IEEE) received the master’s degree from National University of Singapore, and the Ph.D. degree from Monash University. She has about 6 years of working experience at the enterprise. She is currently a professor at the School of Information Science and Technology, Harbin Institute of Technology (Shenzhen), China. Her research interests are multimedia computing and information retrieval. She has published more than 60 papers at the first-tier conferences and journals, like ACM MM, SIGIR, IEEE TPAMI and IEEE TIP.
\end{IEEEbiography}

\begin{IEEEbiography}[{\includegraphics[width=1in,height=1.25in,clip,keepaspectratio]{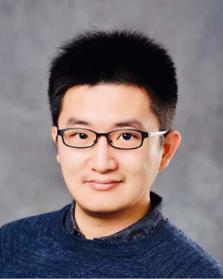}}]
{Xiangyu Zhao} (Member, IEEE) received the B.E. degree from UESTC, in 2014, the M.S. degree from USTC, in 2017, and the Ph.D. degree from MSU, in 2021. He is a tenured associate professor with the Department of Data Science, City University of Hong Kong (CityU). His research has been awarded ICDM’22 and ICDM’21 Best-ranked Papers and Global Top 25 Chinese New Stars in AI (Data Mining).
\end{IEEEbiography}

\begin{IEEEbiography}[{\includegraphics[width=1in,height=1.25in,clip,keepaspectratio]{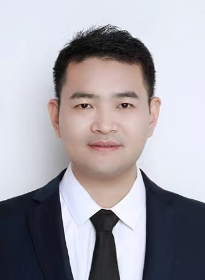}}]
{Liqiang Nie} (Senior Member, IEEE) is currently the dean with the Department of Computer Science and Technology, Harbin Institute of Technology (Shenzhen). He received his B.Eng. and Ph.D. degree from Xi'an Jiaotong University and National University of Singapore (NUS), respectively. After PhD, Dr. Nie continued his research in NUS as a research fellow for three years. His research interests lie primarily in multimedia computing and information retrieval. Dr. Nie has co-authored more than 100 papers and 4 books, received more than 41,000 Google Scholar citations. He is an AE of IEEE TKDE, IEEE TCSVT, ACM ToMM, Information Science. Meanwhile, he is the regular area chair of ACM MM, NeurIPS, IJCAI and AAAI. He is a member of ICME steering committee. He has received many awards, like ACM MM and SIGIR best paper honorable mention in 2019, SIGMM rising star in 2020, TR35 China 2020, DAMO Academy Young Fellow in 2020, SIGIR best student paper in 2021, ACM MM best paper in 2022.
\end{IEEEbiography}

\end{document}